\pdfoutput=1
\documentclass[11pt]{article}
\usepackage[margin=1in]{geometry}
\usepackage{microtype}
\usepackage{graphicx}
\usepackage{subfigure}
\usepackage{booktabs}
\usepackage{adjustbox}
\usepackage{float}
\usepackage{amsmath,amssymb,amsthm}
\usepackage{tikz}
\usetikzlibrary{decorations.pathreplacing,arrows.meta,positioning}
\usepackage[round]{natbib}
\setcitestyle{authoryear,round,citesep={;},aysep={,},yysep={;}}
\renewcommand{\cite}[1]{\citep{#1}}
\usepackage{authblk}
\usepackage{url}
\usepackage{hyperref}
\definecolor{linkblue}{rgb}{0,0.08,0.45}
\hypersetup{colorlinks=true,linkcolor=linkblue,citecolor=linkblue,urlcolor=linkblue,
  pdftitle={ShiftCache: Fast and Exact Indel Effect Prediction by Reusing Shifted Reference Encodings},
  pdfauthor={Sina Barazandeh, Arth Banka, Jose Lugo-Martinez},
  pdfkeywords={efficient inference, computation reuse, incremental computation, genomic sequence models, variant effect prediction}}
\newcommand{\doi}[1]{doi: \href{https://doi.org/#1}{\nolinkurl{#1}}}

\makeatletter
\g@addto@macro\UrlBreaks{\do\a\do\b\do\c\do\d\do\e\do\f\do\g\do\h\do\i\do\j\do\k\do\l\do\m\do\n\do\o\do\p\do\q\do\r\do\s\do\t\do\u\do\v\do\w\do\x\do\y\do\z\do\A\do\B\do\C\do\D\do\E\do\F\do\G\do\H\do\I\do\J\do\K\do\L\do\M\do\N\do\O\do\P\do\Q\do\R\do\S\do\T\do\U\do\V\do\W\do\X\do\Y\do\Z\do\0\do\1\do\2\do\3\do\4\do\5\do\6\do\7\do\8\do\9}
\makeatother

\newtheorem{theorem}{Theorem}
\newtheorem{lemma}{Lemma}
\newtheorem{corollary}{Corollary}
\newtheorem{proposition}{Proposition}
\newcommand{\restatedname}{}
\newtheorem*{restatedthm}{\restatedname}
\newenvironment{restated}[2]{\renewcommand{\restatedname}{#1}\begin{restatedthm}[#2]}{\end{restatedthm}}
\newenvironment{restatednonote}[1]{\renewcommand{\restatedname}{#1}\begin{restatedthm}}{\end{restatedthm}}
\newcommand{\dedupdense}{Dedup dense}
\newcommand{\threepass}{Three-pass dense}

\title{ShiftCache: Fast and Exact Indel Effect Prediction\\ by Reusing Shifted Reference Encodings}
\author[1]{Sina Barazandeh}
\author[1,2]{Arth Banka}
\author[1]{Jose Lugo-Martinez\thanks{Corresponding author: \texttt{jlugomar@andrew.cmu.edu}}}
\affil[1]{Ray and Stephanie Lane Computational Biology Department, Kenneth C.\ Griffin School of Computer Science, Carnegie Mellon University, Pittsburgh, PA, USA}
\affil[2]{Department of Immunology, University of Pittsburgh School of Medicine, Pittsburgh, PA, USA}
\date{}

\begin{document}
\maketitle

\begin{abstract}
Sequence-to-function models such as AlphaGenome and Borzoi read up to a megabase of DNA and predict thousands of genomic signals. A genetic variant is scored by comparing a model's predictions for the reference and alternate sequences. Unlike single nucleotide variants, insertions and deletions (indels) are costly to score, because an indel usually shifts the unchanged DNA after it onto a different pooling grid, which prevents direct reuse of the reference's encoder activations. The AlphaGenome Atlas, for example, scored more than 100 million indels with three full model passes each, and every pass encodes the entire window, although the DNA after an indel matches the reference shifted by the indel's length. To avoid this repeated work, we present ShiftCache, an exact reuse system that encodes each shifted reference once and shares it among all variants scored in the same window with the same shift. A planner derives which encoder outputs each variant can copy, and ShiftCache recomputes only the rest in batched tiles. On held-out AlphaGenome windows shared by many indels, in strict fp32 on one A6000, ShiftCache is 3.5 to 6.5$\times$ faster than deduplicated dense execution and 1.6 to 2.7$\times$ faster than our adaptation of layer-by-layer reuse to indels. It is 3.5 to 4.1$\times$ faster than dense reimplementations of two published analyses with Borzoi and AlphaGenome, and it speeds up Enformer scoring by 2.6$\times$ over dense execution. Scoring each variant in its own centered 131~kb window makes ShiftCache only 1.36$\times$ faster than deduplicated dense execution, and highly diverse insertion lengths or many dispersed edits can make it slower than layer-by-layer reuse or dense execution. For encoders that meet our conditions, ShiftCache computes the same predictions as dense execution in exact arithmetic. On an A6000, every checked output of the default AlphaGenome and Borzoi runs matches dense execution within an absolute and relative tolerance of $10^{-5}$. With larger tile batches, with Enformer or on an H100, a small fraction of values exceeds this tolerance.

\end{abstract}

\section{Introduction}
\label{sec:intro}

Predicting the effect of a genetic variant is a common use of genomic sequence models. The model runs on the
reference sequence and on the edited sequence (the \emph{alternate}), and the two predictions are compared. Models such as
AlphaGenome~\cite{avsec2026alphagenome} and Borzoi~\cite{linder2025borzoi} read up to a megabase of DNA. Most of
their computation occurs in a convolutional encoder that precedes a transformer, and at 131~kb AlphaGenome spends about
90\% of a forward pass in that encoder (\S\ref{sec:breakdown}). An edit changes only a few bases, so most of the alternate's encoding should be reusable from the reference. That holds for substitutions, but for insertions and deletions
(\emph{indels}) reuse is harder.

The AlphaGenome Atlas, a genome-wide catalog of predicted variant effects, scores more than 100 million
indels~\cite{alphagenome_atlas}, and Borzoi-based fine-mapping, which ranks candidate causal variants, scores about
1.9 million~\cite{srivastava2025finemapping}.
Besides the reference and the alternate, indel \emph{stitching}~\cite{korsakova2025shift,alphagenome_research} evaluates a third sequence so that
predictions line up on both sides of the edit. For AlphaGenome, this third sequence is a shifted reference for a deletion and a second alternate for an insertion (\S\ref{sec:problem}), so the Atlas runs three full passes per indel. Each pass encodes the entire window, although the three sequences differ only near the indel and by a shift.

This repeated work is hard to avoid because the encoder pools its input. AlphaGenome's encoder reduces the input to one output position (\emph{bin}) per 128 bases with seven stride-2 pools. An indel shifts everything after it, and unless the shift is a multiple of 128, the shifted sequence falls on a different pooling grid. If the indel length is odd, for example, the first stride-2 pool pairs bases differently than in the reference, so even the unchanged sequence after the indel can produce different
encoder outputs. The reference's outputs therefore cannot simply be moved by the shift and copied.

Existing reuse methods eliminate only part of this work for indels. Caching outputs per input sequence, which we call deduplicated dense execution, skips sequences that repeat across requests but still encodes every distinct alternate in full. Prefix caching in LLM serving~\cite{zheng2024sglang} reuses only the input before the edit, so it recomputes everything after the edit, which is half the window for a centered indel. Layer-by-layer reuse for in silico mutagenesis, which scores every possible
substitution~\cite{nair2022fastism}, can reuse the reference's activations, moved by the shift, only in layers whose stride divides the shift. We call its adaptation to indels layer-local reuse. Among the 1,953 real 1000 Genomes indels in our held-out 131~kb windows, 57\% have odd length. Counting only the pooling levels whose stride divides each indel's length, layer-local reuse can avoid at most about 40\% of the convolution work after the edit, averaged over these indels.

In this work, we present ShiftCache, a system that restores the pooling alignment by encoding a \emph{shifted reference}, the reference moved by the indel's length. After the indel, the alternate matches this shifted reference base for base and lies on the same pooling grid, so the alternate can copy from it every encoder output that depends only on that region. Indels scored in the same window often share a shift, and one shifted reference then serves all of them. A scan of 512 deletions of 1 to 4 bases in one window, for example, needs only four shifted references. The stitched AlphaGenome passes for deletions already include these references, so sharing them adds no encoder pass.

ShiftCache processes a \emph{campaign}, a set of variant-scoring requests handled together, and encodes each shared shifted reference once as a \emph{view}. A planner uses the encoder's kernel sizes and pooling strides to compute which bins each request can copy from the reference or a view. ShiftCache recomputes the remaining bins in batched input tiles and runs the transformer and prediction heads unchanged. Its outputs equal those of dense execution in exact arithmetic. ShiftCache builds on local recomputation for edits that keep positions fixed, as in fastISM, and on evaluating one input at every alignment of the pooling grid (\S\ref{sec:background}). It extends these ideas by sharing shifted references across requests.

Sharing helps most when many indels can share one window, as in analyses that group nearby variants~\cite{baskerville}, deletion scans of promoters (the regulatory DNA at the start of a gene)~\cite{korsakova2025shift}, and deletions of candidate regulatory sequences and length-matched
controls~\cite{motifablation2026}. On shared-window AlphaGenome campaigns, ShiftCache is 3.5 to 6.5$\times$ faster than deduplicated dense execution and 1.6 to 2.7$\times$ faster than layer-local reuse adapted to indels. On two reimplemented published analyses with Borzoi and AlphaGenome, it is 3.5 to 4.1$\times$ faster than dense runs of these reimplementations. When each indel is scored in its own centered window, as in AlphaGenome's tutorial, less work is
shared, and ShiftCache is 1.36$\times$ faster than deduplicated dense execution at 131~kb (\S\ref{sec:limits}).

We make three contributions.
\begin{enumerate}
\item \textbf{Shifted-reference reuse and ShiftCache.} We introduce a mechanism and executor for exact computation reuse across requests whose inputs contain
  shifted segments of a shared reference, with a closed-form planner for one or several encoder outputs
  (\S\ref{sec:design}).
\item \textbf{Correctness guarantees and validation.} We prove that every bin ShiftCache copies or recomputes equals the dense encoder output in exact arithmetic (\S\ref{sec:correctness}), and we compare full fp32 outputs with dense execution on an RTX A6000 and an H100 (\S\ref{sec:exact}). The small differences we find likely come from cuDNN's shape-dependent choice of convolution algorithm. With a fixed-order convolution kernel, the checked AlphaGenome encoder outputs are bitwise identical to dense execution using the same
  kernel (\S\ref{sec:execution}, \S\ref{sec:exact}).
\item \textbf{Performance and limitations analysis.} We evaluate ShiftCache on AlphaGenome, Borzoi and Enformer, including two reimplemented published analyses, and show when it is slower than layer-local reuse or dense execution (\S\ref{sec:results}).

\end{enumerate}

\section{Background}
\label{sec:background}

We introduce the three models we accelerate and the prior work that reuses computation within one input.

\paragraph{Genomic sequence models.} These models map one-hot encoded DNA to genomic \emph{tracks}, and each track is a predicted
measurement along the sequence, such as gene expression in one cell type. A \emph{bin} is one output position, and a \emph{head} predicts one type of measurement. AlphaGenome~\cite{avsec2026alphagenome} reads up to $2^{20}$ bases, and its
encoder uses odd-width convolutions with length-preserving zero padding,
normalization with stored statistics, residual additions and seven kernel-2, stride-2 max pools. The encoder output,
which we call the \emph{trunk}, has 1,536 channels per 128-base bin and feeds a transformer with pairwise attention (the \emph{tail}) and then the heads. Borzoi~\cite{linder2025borzoi} reads 524,288 bases, which a convolutional encoder pools to 128-base resolution before a transformer. A U-Net then upsamples using encoder activations at 32- and 64-base resolution, and Borzoi predicts 7,611 tracks at 32-base resolution over the central 196,608 bases. Enformer~\cite{avsec2021enformer}
reads 196,608 bases and downsamples by attention pooling, a softmax-weighted sum over aligned pairs with learned weights.

\paragraph{Local computation reuse.} Each output of a convolutional encoder depends only on a fixed interval of its input, its receptive field. fastISM~\cite{nair2022fastism} uses this locality to accelerate in silico mutagenesis by
recomputing, at each layer, only the activations that depend on the substituted base.
Krypton~\cite{nakandala2019krypton} explains CNN predictions by masking same-size image patches, treats each masked query
as an incremental update to the CNN's activations, and batches queries. Both handle edits that keep positions fixed, so neither reuses work across the shift an indel causes. Computing each aligned tile with a halo of extra input makes tile outputs identical to full-input outputs~\cite{majurski2019tiling}, and ShiftCache uses this to recompute the outputs it cannot copy.

\paragraph{Pooling phases.} A \emph{pooling phase} is an alignment of the pooling grid relative to the input.
\citet{giusti2013fast}, \citet{bailer2017fast} and \citet{thom2017rapid} compute the pooled outputs of one input at
every pooling phase to obtain exact sliding-window outputs at all positions. These methods share work within one input, and ShiftCache's optional phase views
(\S\ref{sec:planning}) apply the same idea across requests that share a reference. Shift-and-stitch~\cite{long2015fcn,lin2016mustcnn}, unrelated to indel stitching, interleaves the outputs of
shifted copies of one input.

\section{Problem Formulation}
\label{sec:problem}

This section defines dependency intervals, stitched scoring, campaigns and dense baselines, and then states the problem.

\paragraph{Dependency intervals.} Bin $j$ of AlphaGenome's trunk depends only on input positions in its receptive field,
or \emph{dependency interval}, $[128j-513,\,128j+640]$, clipped to the window. A convolution of radius $r$ applied at cumulative stride $s$, the spacing in input bases between adjacent
positions, widens the interval by $rs$ on both sides, and a pool whose input has stride $s$ widens it by $s$ on the
right. Gradient tests with random weights (\S\ref{sec:eval}) confirm
this interval. Borzoi has three encoder output tensors, at 128-, 64- and 32-base resolution, each with its own
dependency interval.

\paragraph{Stitched indel scoring.} The alternate can be aligned with the reference at the window's start
(left-anchored, $\mathrm{alt}$) or end (right-anchored, $\mathrm{alt}_{\rightarrow}$), and each lines up with the reference on only one side of the edit. Let $H$ be the encoder and transformer, which produce the 128-base
embeddings. For an insertion or deletion of $k\ge1$ bases (forward strand, 128-base resolution), the AlphaGenome
protocol~\cite{korsakova2025shift,alphagenome_research} computes:
\begin{itemize}
\item deletion: reference side $=\mathrm{stitch}(H(\mathrm{ref}), H(S_{-k}))$, alternate side $=H(\mathrm{alt})$;
\item insertion: reference side $=H(\mathrm{ref})$, alternate side $=\mathrm{stitch}(H(\mathrm{alt}), H(\mathrm{alt}_{\rightarrow}))$,
\end{itemize}
where $S_{-k}$ is the reference shifted left by $k$, which we call \emph{right-matched} because it matches the
alternate after the deletion. Stitching joins two sets of embeddings at a bin determined by the edit's endpoints, taking earlier bins from the first and later bins, shifted by a whole number of bins, from the second, and the heads then run on the stitched embeddings. Borzoi's scoring library, baskerville~\cite{baskerville}, instead stitches 32-base predictions, and only on the alternate side, where
it combines both anchored alternates. Our baseline \threepass{} runs three sequences per indel. For AlphaGenome, a deletion needs the reference, $S_{-k}$ and the alternate,
and an insertion needs the reference and both anchored alternates. The Atlas
caches references for substitutions but not for indels~\cite{alphagenome_atlas}, and baskerville predicts the reference once per cluster of nearby variants~\cite{baskerville}.

\paragraph{Campaigns and dense baselines.} In a \emph{shared-window campaign}, all requests use the same reference
window. Under AlphaGenome's stitched protocol, such a campaign needs the reference, one right-matched reference $S_{-k}$ per deletion length, one alternate per deletion and two anchored alternates per insertion, so many
requests share sequences. The strongest dense baseline, deduplicated dense execution (\dedupdense{}), runs each
distinct sequence once. It shares the reference and the right-matched references but still encodes every alternate in
full. In AlphaGenome's quick-start tutorial, each indel is scored in its own centered window. Requests can then share work only within one indel's passes and across nearby indels' windows, which are shifted copies of
one another.

\paragraph{Problem statement.} The inputs are a reference window, a campaign of edits, a pretrained model whose encoder meets the locality conditions of \S\ref{sec:correctness}, and a scoring protocol. The window comes with its \emph{flanks}, the neighboring genomic bases that its shifts bring into the window. The goal is to compute every request's reference and alternate outputs with less encoder work than \dedupdense{}. The assembled encoder outputs,
and therefore the scores, must equal those of dense execution in exact arithmetic. We evaluate floating-point agreement
separately (\S\ref{sec:exact}).

\section{ShiftCache Design}
\label{sec:design}

\begin{figure}[t]
\centering
\begin{tikzpicture}[x=0.29cm,y=0.5cm,font=\scriptsize,
  bin/.style={draw=white,line width=0.6pt,minimum width=0.29cm,minimum height=0.42cm,inner sep=0},
  lab/.style={anchor=east}]
  \foreach \i in {0,...,19} {\fill[gray!35] (\i,3) rectangle (\i+1,3.8);}
  \node[lab] at (-0.2,3.4) {$E(S_0)$};
  \foreach \i in {0,...,19} {\fill[orange!55] (\i,2) rectangle (\i+1,2.8);}
  \node[lab] at (-0.2,2.4) {$E(S_{-k})$};
  \fill[gray!35] (0,0.6) rectangle (8,1.4);
  \fill[red!65] (8,0.6) rectangle (12,1.4);
  \fill[orange!55] (12,0.6) rectangle (20,1.4);
  \node[lab] at (-0.2,1.0) {$E(\mathrm{alt})$};
  \foreach \i in {1,...,19} {\draw[white,line width=0.6pt] (\i,0.6) -- (\i,1.4);
    \draw[white,line width=0.6pt] (\i,2) -- (\i,2.8); \draw[white,line width=0.6pt] (\i,3) -- (\i,3.8);}
  \draw[-{Stealth[length=3pt]},gray!70!black] (4,3) -- (4,1.45);
  \draw[-{Stealth[length=3pt]},orange!80!black] (16,2) -- (16,1.45);
  \draw[decorate,decoration={brace,mirror,amplitude=3pt}] (8,0.5) -- (12,0.5);
  \node[anchor=north] at (10,0.25) {recomputed};
  \node[anchor=north] at (4,0.25) {from $E(S_0)$};
  \node[anchor=north] at (16,0.25) {from $E(S_{-k})$};
  \draw[dashed,red!70!black] (10,0.6) -- (10,4.05) node[above] {deletion of $k$ bases};
  \fill[gray!35] (0,-1.3) rectangle (2,-0.5);
  \fill[red!65] (2,-1.3) rectangle (6,-0.5);
  \fill[orange!55] (6,-1.3) rectangle (20,-0.5);
  \foreach \i in {1,...,19} {\draw[white,line width=0.6pt] (\i,-1.3) -- (\i,-0.5);}
  \node[lab] at (-0.2,-0.9) {$E(\mathrm{alt}')$};
  \draw[-{Stealth[length=3pt]},orange!80!black] (20.1,2.4) to[out=0,in=0] (20.1,-0.9);
  \node[anchor=north,font=\scriptsize\itshape] at (13,-1.35) {same view, another deletion of $k$ bases};
  \begin{scope}[shift={(1.5,-3.3)}]
    \node[anchor=west,font=\scriptsize\itshape] at (-1.7,0.55) {Why: the first pool pairs positions $(0,1),(2,3),\dots$};
    \foreach \row/\y/\seq/\pc in {{$S_0$}/-0.35/{A,B,C,D,E,F,G,H}/gray!70,
                                   {alt}/-1.25/{A,C,D,E,F,G,H,I}/orange!80!black,
                                   {$S_{-1}$}/-2.15/{B,C,D,E,F,G,H,I}/orange!80!black} {
      \node[lab] at (-0.4,\y) {\row};
      \foreach \b [count=\n from 0] in \seq {\node at (1.5*\n+0.75,\y) {\texttt{\b}};}
      \foreach \j in {0,...,3} {\draw[rounded corners=2pt,\pc] (3*\j+0.1,\y-0.3) rectangle (3*\j+2.9,\y+0.3);}
    }
    \draw[red!70!black,thick] (1.75,-0.6) -- (2.75,-0.1);
    \node[anchor=west] at (12.2,-0.35) {pairs \texttt{CD}, \texttt{EF}};
    \node[anchor=west] at (12.2,-1.25) {pairs \texttt{DE}, \texttt{FG}};
    \node[anchor=west] at (12.2,-2.15) {same pairs: copy};
  \end{scope}
\end{tikzpicture}
\caption{One shifted encoding serves many requests. To assemble a deletion's left-anchored alternate encoding,
bins whose dependency intervals lie before the deletion are copied from the reference, and bins whose intervals lie after it are copied from the reference shifted left by $k$. This shifted encoding is
built once and shared by every request with the same offset, such as a deletion of $k$ bases elsewhere in the window
(row $E(\mathrm{alt}')$), and also serves as the stitching protocol's right-matched reference.
For AlphaGenome, about nine bins have dependency intervals that cross the edit and are recomputed in stride-aligned tiles with a five-bin halo. The inset shows that deleting an input base changes the pooling groups, while $S_{-k}$ matches the alternate's
groups. Copying an output bin requires its entire dependency interval to match.}
\label{fig:mechanism}
\end{figure}
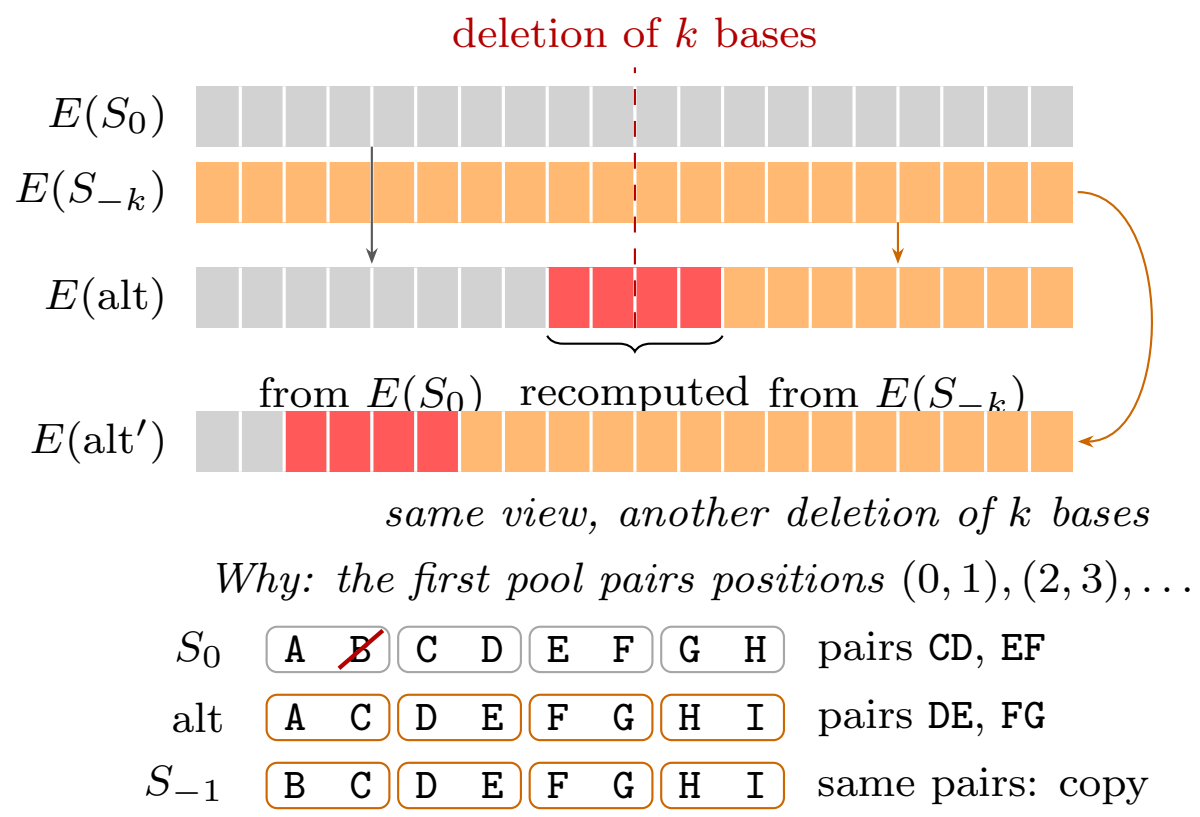

ShiftCache reuses encoder outputs across requests. Because each encoder output depends only on a fixed input interval (\S\ref{sec:problem}), an output bin of an alternate can be copied from any encoding whose input matches over that interval. Let $L$ be the window length
and $E$ the encoder. Each sequence in a campaign consists of reference segments copied
at some offset, possibly with inserted bases. ShiftCache maps these segments to reference coordinates and computes the
reusable bins in closed form. It then encodes the shifted references the campaign needs as shared views and assembles each request's encoder output from copied bins and small recomputed tiles (Figure~\ref{fig:mechanism}). The transformer (the tail) and heads run unchanged.

\subsection{Reuse Planning from Shifted References}
\label{sec:planning}

\paragraph{Segment representation.} Applying edits to a reference window with real genomic flanks yields a sequence $X$ and
segments $[a,e)$ with offsets $d$ such that $X[i]=\mathrm{ref}[i-d]$ for $i\in[a,e)$. A left-anchored deletion of
$k$ bases at $p$ has segments $[0,p)$ with $d=0$ and $[p,L)$ with $d=-k$, whose last $k$ bases come from the right
flank. Right-anchored deletion alternates keep the sequence after the edit in place and take their extra bases from the
left flank. References, right-matched references, alternates and
centered windows are all described this way.

\paragraph{Clean bins in closed form.} Let an encoder output tensor have stride $s$ and
dependency interval $[sj+\ell,\,sj+h]$ for index $j$. For AlphaGenome's trunk, $s{=}128$ and $(\ell,h){=}(-513,640)$, and the total encoder stride is $P{=}128$.
Bin $j$ is \emph{clean with source $d$} if its interval, clipped to $[0,L)$, lies inside one segment $[a,e)$ with offset $d$.
Among valid output indices, the clean bins satisfy
\begin{equation}
\begin{split}
j &\ge \lceil(a-\ell)/s\rceil \quad \text{if }a>0,\\
j &\le \lfloor(e-1-h)/s\rfloor \quad \text{if }e<L.
\end{split}
\end{equation}
A clean bin equals bin $j$ of
$E(S_d)$, the encoding of the reference shifted by $d$, and both are padded with zeros in the same way at the window edges
(\S\ref{sec:correctness}). The remaining \emph{dirty} bins have dependency intervals that cross an edit or include
inserted bases. With the required views built, a short AlphaGenome edit leaves about nine dirty bins.
Clean bins whose source view is not built are also recomputed. The planner represents each request as a run-length list of
(bin range, source) pairs.

\paragraph{Shared views.} A view is an encoding $E(S_d)$ shared by the requests that need offset $d$.
Under AlphaGenome's stitched deletion protocol, the views encode right-matched references that dense execution already evaluates, so sharing them requires no extra encoder pass.
Other views, such as those for insertions, cost extra encoder work and save time only if enough requests share them.

\paragraph{Phase views.} An optional mode keeps one view per pooling phase and reindexes it for every offset with that phase. An offset $d=\phi+Pq$ with phase $0\le\phi<P$ gives, for the trunk, $E(S_d)[j]=E(S_\phi)[j-q]$
wherever both dependency intervals lie inside their windows, so at most $P$ views are needed and boundary bins are
recomputed. The evaluated default keeps one view per offset.

\subsection{Tiled Execution and View Selection}
\label{sec:execution}

\paragraph{Tiles and assembly.} Bins that cannot be copied are recomputed from input tiles whose start
and length are multiples of $P$. Each tile
carries a halo on each side whose outputs are discarded, so that every retained bin sees
the same input as in the full window. For AlphaGenome, each tile spans 32 or 128 trunk bins, including a five-bin halo on each
side, so an interior tile keeps 22 or 118 bins. Tiles are batched across requests, up to 16 per batch (the \emph{tile batch}). Batch sizes are rounded up to powers of two to limit the number of compiled shapes. A request whose tiles would cost more than a full encode falls back to a full pass. Assembly
gathers bins from the reference, the views and the tile outputs. For Borzoi, the planner combines the dependency
intervals of all three encoder outputs on the 128-base trunk grid, so one plan and tile layout cover them all.

\paragraph{View selection.} A full-window view requires one encoder pass, while a partial view covers only
the smallest interval containing the bins copied from it and is built from stride-aligned tiles.
The static view-selection rule, used by default, builds a view for every offset used by at least two
distinct sequences. The greedy rule adds a view only when doing so lowers the predicted campaign time, using a cost model based on GPU measurements of per-batch costs for full encoder passes and tiles (the \emph{cost table}). Choosing which encodings to keep is a form of
materialized-view selection~\cite{harinarayan1996cubes}. With many dispersed edits, the greedy rule avoids views that save little work (\S\ref{sec:limits}).

\paragraph{Campaign execution.} The executor deduplicates the campaign's sequences, creates a reuse plan, builds
the views once and assembles encoder outputs, in chunks of requests for large campaigns. It then runs the unchanged downstream model and
stitches at the stage the protocol prescribes (\S\ref{sec:problem}). Each distinct sequence's embeddings are computed when first needed and kept until their last use.

\paragraph{Bitwise mode.} cuDNN can choose different summation orders for different tensor shapes, so a bin computed
in a tile may differ in its last bits from the same bin computed in a full window. An optional mode routes every
AlphaGenome encoder convolution, in all execution methods including dense execution, through one fixed-order Triton
kernel (Appendix~\ref{app:bitwise}). The kernel performs the same operations at each output position regardless of
tensor length, tile offset or batch size. We evaluate this mode only for AlphaGenome under \texttt{torch.compile} (\S\ref{sec:exact}).

\subsection{Correctness Guarantees}
\label{sec:correctness}

Assembly is valid only if every copied bin has matching input dependencies and every recomputed tile keeps the pooling
alignment of the full window. Both requirements hold under the following conditions.

\paragraph{Encoder assumptions.} Let $E$ be a strided encoder built from stride-1, dilation-1 convolutions with odd kernels and symmetric, length-preserving zero
padding, spatially shared deterministic pointwise operations, residual additions, normalization with frozen
statistics, and pooling. Each pooling layer applies the same deterministic function, with shared parameters, to
aligned non-overlapping pairs $(2j,2j{+}1)$, as AlphaGenome's and Borzoi's max pools do, and no operation depends on absolute position, input length or batch
contents. The input has length $L$, a multiple of the total stride $P=2^p$, where $p$ is the number of pools, with
zero padding outside $[0,L)$. For encoder output tensor $b$ with stride $s_b$, the structural dependency interval of index $j$
is $I_b(j)=[s_bj+\ell_b,\,s_bj+h_b]$, computed as in \S\ref{sec:problem}, and its clipped interval is
$\bar I_b(j)=I_b(j)\cap[0,L)$.

\begin{lemma}[Locality]
$E(X)_b[j]$ depends only on $X$ restricted to $\bar I_b(j)$ and on which ends of $I_b(j)$ were clipped.
\end{lemma}

\begin{theorem}[Copy validity]
\label{thm:copy}
Let $Y$ be a reference sequence defined on the window and the genomic flanks needed by the shifts, and let
$Y_d[i]=Y[i-d]$ for $0\le i<L$. If $X[i]=Y_d[i]$ for all $i\in\bar I_b(j)$, then $E(X)_b[j]=E(Y_d)_b[j]$ (both are
padded identically at the window edges).
\end{theorem}
For $Y=\mathrm{ref}$, $Y_d$ is the shifted reference $S_d$ of \S\ref{sec:design}.

\begin{theorem}[Phase reindexing]
\label{thm:phase}
If $d=\phi+Pq$ with integer $q$ and $0\le\phi<P$, then $E(Y_d)_b[j]=E(Y_\phi)_b[j-(P/s_b)q]$ wherever both intervals are unclipped. For shifts not divisible by
$2^l$, the assumptions do not in general guarantee any integer reindexing of the unshifted encoding at stride $2^l$,
because the pooling groups differ, although particular inputs or weights can yield coincidentally equal outputs.
\end{theorem}

\begin{theorem}[Tile exactness]
\label{thm:tile}
Let $X_t=X[tP,\,tP+T)$ be a tile contained in the window, with integer $t\ge0$ and $T$ a positive multiple of $P$.
Then $E(X_t)_b[j-tP/s_b]=E(X)_b[j]$ whenever $\bar I_b(j)$ lies inside the tile, with
$tP=0$ if $s_bj+\ell_b<0$ and $tP+T=L$ if $s_bj+h_b\ge L$. Thus, a dependency interval may cross a tile
boundary only where that boundary coincides with the corresponding window edge.
\end{theorem}

\begin{corollary}
In exact arithmetic, assembling every bin according to Theorems~\ref{thm:copy} to~\ref{thm:tile} yields $E(X)$.
Applying the same tail, stitching and heads then gives the dense outputs.
\end{corollary}

\noindent Appendix~\ref{app:proofs} gives full proofs of these results and derives the clean-bin conditions of
\S\ref{sec:planning}.

\section{Evaluation Methodology}
\label{sec:eval}

We evaluate ShiftCache on held-out windows with real indels, deletion and insertion scans, and two reimplemented published analyses. We compare it with dense baselines and layer-local reuse.

\paragraph{Hardware and software.} We use one 48~GB NVIDIA RTX A6000 and PyTorch 2.6 with CUDA
11.8~\cite{paszke2019pytorch}. Every execution method compiles its encoder and transformer with \texttt{torch.compile}~\cite{ansel2024pytorch2} and cuDNN
autotuning, and we run strict fp32 with TF32 disabled unless stated otherwise. We use the AlphaGenome PyTorch
port~\cite{alphagenome_pytorch} with the released checkpoint trained on all data folds, the Borzoi PyTorch
port~\cite{borzoi_pytorch}, and the pretrained Enformer PyTorch port~\cite{enformer_pytorch}. We repeat key runs on an
NVIDIA H100 with the same pinned packages, weights and inputs (Appendix~\ref{app:repro}).

\paragraph{Genomic windows and workloads.} Design and tuning windows lie on chromosomes 1, 17 and
19. After tuning ShiftCache, we froze its settings and drew four held-out windows on chromosomes 2, 6, 11 and 12. We first evaluated them with that code, and all later held-out runs use these windows. The layer-local baseline was added and improved after that evaluation (Appendix~\ref{app:protocol}). Each 524,288-base window contains a centered 131,072-base window, and both have real genomic flanks (Appendix~\ref{app:protocol}). We use four workloads. \emph{Real indels} are sampled with a fixed seed from the indels of at most 50 bases observed in the 3,202 high-coverage genomes of the 1000
Genomes Project~\cite{byrska2022kgp}, with 256 variants per 131~kb window and 128 per 524~kb window. A
\emph{deletion scan} covers every deletion of 1 to 4 bases starting in a central region of 128 bases at 131~kb or 64 at 524~kb,
giving 512 and 256 requests.
\emph{Matched insertions} are 128 (131~kb) or 64 (524~kb) distinct random 12-base insertions at one position, like the length-matched background insertions
of AlphaGenome's TAL1 analysis~\cite{avsec2026alphagenome}. A separate \emph{exhaustive scan}, on two held-out 131~kb windows, covers every deletion and every possible insertion
of 1 to 4 bases at 32 sites, giving $32\times(4+4+16+64+256)=11{,}008$ requests. These campaigns use AlphaGenome's gene-expression (RNA-seq) head at 128-base resolution on the forward strand and
its stitching protocol (\S\ref{sec:problem}). We discuss its official gene-level scores in \S\ref{sec:limitation}. \emph{Unstitched} runs instead score one left-anchored
alternate per variant.

\paragraph{Published-analysis protocols.} We reimplement two published analyses. \citet{korsakova2025shift} score a single-base deletion at every base of the HBG1 and LDLR promoters with Borzoi and reference stitching, at three passes per base. We evaluate the same coordinates with the first of Borzoi's four replicates in one 524~kb window centered on each promoter, and compare stitched profiles of all 7,611 tracks over the central 4~kb, the region they score. Our adapter instead follows baskerville and stitches the alternate, which needs up to $2n{+}1$ sequences for $n$ single-base deletions in one window, against $n{+}2$. Because the published analysis also uses a four-replicate ensemble in per-variant centered windows, our reimplementation does not reproduce its exact figures. \citet{motifablation2026} test 11 binding sites (motifs) of the transcription factor CEBPB near the
TSC22D3 gene. For each motif, the analysis deletes the 12-base motif and $K$ length-matched control segments drawn uniformly from
a 1~Mb window centered on the motif. It then predicts ChIP-seq tracks of histone marks (chemical tags on the proteins that package DNA) in the MM.1S myeloma cell line, at 128-base resolution and without stitching. A
motif is classified as significant when its local fold change, the ratio of alternate to reference signal near the motif, falls below the 2.5th or above
the 97.5th percentile of the control fold changes. We follow the published window placement, seeded sampling and significance criteria, but do not time DeepMind's hosted API. At $K{=}3000$, dense execution runs the first 64 deletions per motif to check agreement and compare times.

\paragraph{Baselines and configurations.} \dedupdense{} and \threepass{} are defined in \S\ref{sec:problem}.
Tuning \dedupdense{}'s batch size, stitching chunk size and compilation settings speeds it up by at most 3\% (Appendix~\ref{app:tuning}). \emph{Aligned tiles} is ShiftCache without views. It copies bins only from the
reference trunk, reindexing shifts divisible by 128, and recomputes the rest in the same batched tiles. \emph{Layer-local} is our implementation of fastISM-style layer-by-layer reuse for indels. It caches the reference's convolution outputs before every pool. For a shifted segment, it reuses them up to the
deepest level whose stride divides the shift (only the first, full-resolution level for an odd shift) and
recomputes from that level's pool upward with the same batched tiles. \emph{ShiftCache} is our executor with a separate view per offset and the static view-selection rule. The greedy rule is no faster here (Appendix~\ref{app:ablation}). All methods share the tail,
stitching and heads.

\paragraph{Timing and memory measurement.} Each timed campaign starts with empty sequence and activation caches, in a \emph{warmed} process that has compiled every shape and measured its cost table. Timing
covers deduplication, planning, view construction, tiles, transformer passes, stitching, heads and delivery of each chunk of
predictions to the host, and excludes the final concatenation of chunks, compilation and cost-table calibration unless stated otherwise. We report the median of three campaigns at 131~kb and two at
524~kb, and Appendix~\ref{app:protocol} lists the exceptions. Unless stated otherwise, times are geometric means over the four held-out windows, and speedups are geometric means of per-window ratios with 95\% $t$-intervals on log ratios across the tested
windows. Memory is peak allocated GPU tensor memory.

\paragraph{Output validation.} We compare each method's reference and alternate outputs with a dense baseline. An output value
passes \emph{the tolerance} if it differs from dense execution by at most $10^{-5}+10^{-5}|\mathrm{dense}|$. The campaigns of Table~\ref{tab:main} compare every predicted value, and Borzoi's held-out runs compare each track's sum over positions. CPU tests with random weights check that the input gradient of an interior position is
nonzero up to the ends of its dependency interval and zero beyond them, in every encoder output tensor of AlphaGenome and Borzoi. They also compare the execution methods with dense execution across edit types and both stitching conventions
(Appendix~\ref{app:protocol}).

\section{Results}
\label{sec:results}

We first report speedups, their breakdown and numerical agreement (\S\ref{sec:main} to \S\ref{sec:exact}). We then show the limits of sharing, the transfer to other models and the published analyses (\S\ref{sec:limits} to \S\ref{sec:apps}).

\subsection{Speedup over Dedup Dense and Layer-Local Reuse}
\label{sec:main}

\begin{table}[t]
\centering
\caption{ShiftCache is the fastest method on all six main held-out stitched AlphaGenome workloads (one RTX A6000, strict fp32;
geometric mean seconds over four windows; speedups with 95\% intervals). \threepass{} was timed in an earlier run on the same windows with the frozen code, and all other methods together in a later run.}
\label{tab:main}
\small
\setlength{\tabcolsep}{1.95pt}
\begin{adjustbox}{max width=\textwidth}
\begin{tabular}{llrrrrrrrr}
\toprule
 & & \multicolumn{3}{c}{Time (s)} & \multicolumn{2}{c}{ShiftCache speedup over} & \multicolumn{3}{c}{Other baselines} \\
\cmidrule(lr){3-5}\cmidrule(lr){6-7}\cmidrule(l){8-10}
Context & Workload & \shortstack{Dedup\\dense} & Layer-local & ShiftCache & \shortstack{Dedup\\dense} & Layer-local & \shortstack{Three-pass\\time (s)} & \shortstack{Aligned\\tiles (s)} & \shortstack{Speedup vs.\\three-pass} \\
\midrule
131 kb & 256 real indels & 81.2 & 38.7 & \textbf{18.0} & \textbf{4.52} [4.31, 4.75] & \textbf{2.16} [2.05, 2.26] & 159.0 & 53.7 & 8.9 \\
       & 512-deletion scan & 78.5 & 36.0 & \textbf{13.5} & \textbf{5.81} [5.73, 5.89] & \textbf{2.67} [2.64, 2.69] & 318.0 & 52.7 & 23.5 \\
       & 128 matched insertions & 52.5 & 16.6 & \textbf{8.1} & \textbf{6.47} [6.46, 6.47] & \textbf{2.05} [2.05, 2.05] & 79.5 & 35.5 & 9.8 \\
\midrule
524 kb & 128 real indels & 179.9 & 87.4 & \textbf{50.8} & \textbf{3.54} [3.40, 3.69] & \textbf{1.72} [1.65, 1.79] & 345.3 & 119.6 & 6.8 \\
       & 256-deletion scan & 169.1 & 77.6 & \textbf{36.0} & \textbf{4.70} [4.55, 4.84] & \textbf{2.15} [2.10, 2.21] & 690.5 & 110.1 & 19.2 \\
       & 64 matched insertions & 116.8 & 37.2 & \textbf{22.7} & \textbf{5.15} [5.14, 5.15] & \textbf{1.64} [1.64, 1.64] & 172.6 & 74.6 & 7.6 \\
\bottomrule
\end{tabular}
\end{adjustbox}
\end{table}

\begin{figure}[t]
\centering
\includegraphics[width=0.6\textwidth]{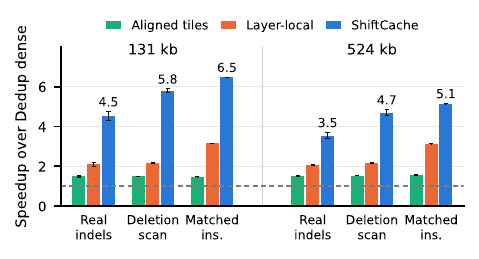}
\caption{ShiftCache is 3.5 to 6.5$\times$ faster than \dedupdense{} (dashed line) and 1.6 to 2.7$\times$ faster than
layer-local reuse. Speedup over \dedupdense{} on held-out stitched AlphaGenome campaigns (RTX A6000, strict fp32, geometric mean of four windows, 95\% intervals). Table~\ref{tab:main} lists the times, and Figure~\ref{fig:results} plots them.}
\label{fig:speedup}
\end{figure}

On held-out windows ShiftCache is 4.5 to 6.5$\times$ faster than \dedupdense{} at 131~kb and 3.5 to 5.1$\times$ at 524~kb
(Table~\ref{tab:main} and Figure~\ref{fig:speedup}), and 6.8 to 23.5$\times$ faster than Atlas-style three-pass execution. An earlier run on the same windows with the frozen code gives the same speedups over \dedupdense{} within 1\%. We attribute the smaller gains at 524~kb to the
transformer, which reuse does not accelerate and which accounts for more of each pass at the longer context.

\paragraph{Comparison with layer-local reuse.} Layer-local reuse, the closest alternative, is 2.1 to 3.2$\times$ faster than \dedupdense{}, and ShiftCache is a further
2.0 to 2.7$\times$ faster at 131~kb and 1.6 to 2.2$\times$ at 524~kb. Layer-local reuse helps most for the 12-base
insertions, whose shift is divisible by four so that the first three levels are reused. It helps least for odd shifts, which reuse only the first, consistent with the 40\% bound of \S\ref{sec:intro}. Once a matching view is built, ShiftCache copies
every clean bin after the edit and shares the construction cost across requests. Aligned tiles, which reuse only shifts
divisible by 128, reach about 1.5$\times$ over \dedupdense{}.

Tuning speeds up layer-local reuse by at most 6.4\% among settings whose outputs pass the tolerance. At a tile
batch of 64, where both methods' outputs fail the tolerance, ShiftCache remains 1.60 to 1.98$\times$ faster than layer-local reuse (Appendix~\ref{app:lltune}). These results
compare our implementations and do not bound layer-local reuse in general.

\paragraph{H100 speedups and memory use.} On an NVIDIA H100, ShiftCache's speedups over \dedupdense{} at 131~kb are
within 3\% of those on the A6000 (Appendix~\ref{app:ablation}). In the AlphaGenome campaigns, ShiftCache uses less peak
allocated GPU memory than \dedupdense{}, as does layer-local reuse at 131~kb (Appendix~\ref{app:numerics}).

\subsection{Encoder Reuse and Runtime Breakdown}
\label{sec:breakdown}

\begin{figure}[t]
\centering
\includegraphics[width=0.7\textwidth]{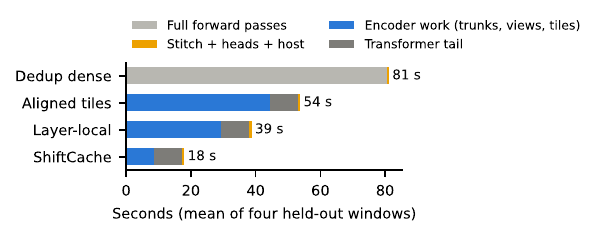}
\caption{Shared views shrink encoder work, and transformer and head time stays the same. Time per campaign by component, AlphaGenome, 256 real indels, 131~kb, one A6000, mean of four held-out windows. \dedupdense's
encoder and tail run together inside its full passes.}
\label{fig:breakdown}
\end{figure}

In the 256-indel campaigns at 131~kb, the transformer alone takes 8.7~s over the 369 to 414 distinct sequences, against 80.5~s for
\dedupdense's full passes, so the encoder accounts for about 90\% of a dense pass
(Figure~\ref{fig:breakdown}). Aligned tiles and layer-local reuse still spend 44~s and 29~s rebuilding encoder outputs,
because most of each alternate lies on a shifted segment they cannot copy. ShiftCache spends 8.5~s on encoder work,
so the transformer accounts for about half of its time. With transformer and output costs unchanged, removing all
encoder work would limit the speedup over \dedupdense{} to about 8.6$\times$. When all required views are available, fewer
than 1\% of alternate trunk bins are dirty (Appendix~\ref{app:ablation}). In deletion scans, every copied bin comes from the reference or from a right-matched reference that
\dedupdense{} computes in full but cannot reuse. In matched insertions, half of the copied bins come from the encodings of $S_{+12}$
and $S_{-12}$, two views that only ShiftCache builds, and in real-indel campaigns 13 to 19\% of copied bins come from such views.

\subsection{Floating-Point Agreement with Dense Execution}
\label{sec:exact}

\paragraph{Agreement on the A6000.} All checked outputs of the held-out campaigns, applications and sensitivity runs pass the tolerance in the default AlphaGenome and Borzoi
configurations. Checks with larger tile batches fail (\S\ref{sec:main}), as do some Enformer outputs
(\S\ref{sec:enformer}). We also checked the full outputs of all 11,008 exhaustive-scan requests on each of its two windows by running each method alongside dense execution, chunk by chunk, and all $1.7\times10^{10}$ values per method and window pass. Full Borzoi
profiles for 41 held-out real indels, covering all 6,144 bins and 7,611 tracks, pass as well.

\paragraph{Agreement on the H100.} On the H100, cuDNN likely sums a 32-bin tile and a full window in different orders. Across the
12 held-out 131~kb campaigns, $5.4\times10^{-6}$ of ShiftCache's output values, in 11\% of requests, exceed the tolerance. Per-campaign median
relative $L_2$ errors in variant effects (alternate minus reference) range from $1.2\times10^{-5}$ to $8.9\times10^{-5}$. In the HBG1 promoter scan with Borzoi, 2,487 of $5.3\times10^8$ values exceed the tolerance, and the LDLR scan has none. Measured against compiled strict-fp32 dense outputs, the errors are of the same order as dense execution's own variation between
batch sizes and between compiled and eager execution (Appendix~\ref{app:numerics}).

\paragraph{Agreement in bitwise mode.} With bitwise mode's fixed-order kernel (\S\ref{sec:execution}), every
AlphaGenome encoder output
we checked on either GPU (34 assembled outputs and 17 views per GPU) is bitwise identical to the output of compiled dense execution with the same kernel. Every
method passes the tolerance in all 12 held-out campaigns. Final outputs differ slightly because operations downstream of the encoder depend on batch size, which changes dense execution's outputs as much. The fixed kernel slows dense execution more than ShiftCache, which
remains 3.15 to 5.49$\times$ faster than default \dedupdense{} on both GPUs (Appendix~\ref{app:bitwise}).

\subsection{Benefits and Limits of Sharing}
\label{sec:limits}

\begin{figure}[t]
\centering
\includegraphics[width=0.6\textwidth]{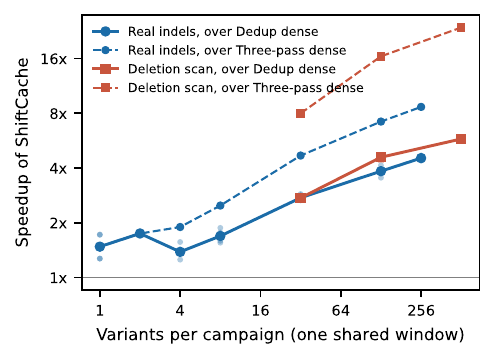}
\caption{Larger campaigns spread view construction over more requests. ShiftCache speedup by campaign size on held-out 131~kb windows (AlphaGenome, one A6000; both axes logarithmic). Solid lines show the
speedup over \dedupdense{} and dashed lines the speedup over \threepass{}; faint points are per-window ratios over
\dedupdense.}
\label{fig:scaling}
\end{figure}

\paragraph{Campaign size and first use.}
\label{sec:scaling}
After warm-up, ShiftCache is faster than \dedupdense{} at every tested campaign size, from 1 to 256 real indels and from 32 to
512 deletions (Figure~\ref{fig:scaling}). ShiftCache is 1.48$\times$ faster even for a single indel (1.27 to 1.72$\times$ across windows), because that indel's passes already share work, as when a deletion's right-matched reference supplies its alternate's suffix. Larger campaigns generally spread the cost of views over more requests, reaching 3.85$\times$ at 128 and 4.5$\times$ at 256 real indels, although small campaigns vary with their composition. The exhaustive scan on two windows has 11,008 requests and about 16,600 distinct
sequences per window but needs only eight views. It runs 7.06$\times$ faster than \dedupdense{}, taking 8 minutes per window against 57. In a fresh process with empty compilation caches, ShiftCache's first campaign takes 194~s with a saved cost table and
347~s when it measures the table, against 222~s for \dedupdense{}. In that case, its cumulative time falls below \dedupdense's by
the fourth campaign (Appendix~\ref{app:tuning}).

\paragraph{Per-variant centered windows.} For 128 real indels per held-out window, each scored in its own centered
131~kb window, ShiftCache is 1.36$\times$ faster than \dedupdense{} (95\% interval 1.34 to 1.38). Requests then share
only one indel's passes and nearby indels' shifted references (\S\ref{sec:problem}).

\begin{figure}[t]
\centering
\includegraphics[width=0.6\textwidth]{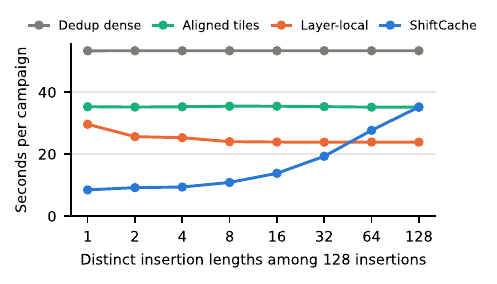}
\caption{Seconds per stitched campaign of 128 random insertions whose lengths take $D$ distinct values (geometric mean
of two held-out windows, 131~kb). Building views stops saving time when too few requests share an offset.}
\label{fig:diversity}
\end{figure}

\paragraph{Offset diversity.} We score 128 random
insertions in each of two held-out 131~kb windows, with lengths that cycle through 1 to $D$ bases, so that each
view serves about $128/D$ requests (Figure~\ref{fig:diversity}). ShiftCache's time grows from 8.4~s at $D{=}1$ to
35.2~s at $D{=}128$, where no offset recurs, so it builds no views and reduces to aligned tiles. Layer-local
reuse needs no sharing and takes 23.8 to 29.6~s, and it is slowest at $D{=}1$, where the single odd length allows reuse
only at the first level. ShiftCache stays faster than layer-local reuse up to $D{=}32$ (19.3~s against 23.8~s). With only one or two requests
per offset, layer-local reuse is 1.16 to 1.48$\times$ faster than ShiftCache, and both remain faster than \dedupdense{} (53~s).
Stitched deletions avoid this regime, because their views are right-matched references that dense execution computes anyway
(\S\ref{sec:planning}).

\begin{table}[t]
\centering
\caption{Runtime in seconds of our implementations of the promoter deletion scans of \citet{korsakova2025shift} and the
TSC22D3 motif-ablation scan~\cite{motifablation2026}, with dense execution and with ShiftCache (RTX A6000, fp32). The Dense protocol column runs three passes per base for the promoter scans, with baskerville's alternate stitching, and one pass per
deletion plus one reference pass for the motif scan. Motif times are per motif, and ranges are across the 11 motifs, each with a 1~Mb window.
\dedupdense{} matches the protocol for the motif scan, whose sequences are nearly all distinct.}
\label{tab:apps}
\footnotesize
\setlength{\tabcolsep}{2.5pt}
\begin{adjustbox}{max width=\textwidth}
\begin{tabular}{lllrrrrr}
\toprule
Analysis & Model, context & Requests & \shortstack{Dense\\protocol} & \shortstack{Dedup\\dense} & ShiftCache & \shortstack{Over\\protocol} & \shortstack{Over\\dedup dense} \\
\midrule
HBG1 promoter scan & Borzoi, 524 kb & 274 & 266.0 & 125.1 & 64.6 & 4.1$\times$ & 1.94$\times$ \\
LDLR promoter scan & Borzoi, 524 kb & 318 & 308.7 & 148.6 & 76.7 & 4.0$\times$ & 1.94$\times$ \\
TSC22D3 motifs, $K{=}500$ & AlphaGenome, 1 Mb & $11\times501$ & 994 to 1,001 & n/a & 265 to 267 & 3.75$\times$ & n/a \\
TSC22D3, $K{=}3000$, first 64 & AlphaGenome, 1 Mb & $11\times64$ & 129 & n/a & 37 & 3.49$\times$ & n/a \\
\bottomrule
\end{tabular}
\end{adjustbox}
\end{table}

\paragraph{Edit density and personal haplotypes.} We run unstitched
campaigns of 32 sequences on two held-out windows, each sequence carrying $m$ random indels of 1 to 4 bases. With the
edits clustered within 4~kb, ShiftCache stays 1.3 to 2.7$\times$ faster than dense execution up to $m{=}64$. With the
edits dispersed across the window, the static rule builds up to 55 views, each shared by few sequences, and ShiftCache becomes
2.0$\times$ slower than dense execution at $m{=}32$. Greedy view selection builds no views there and is at most 7\% slower than dense execution across the dispersed sweep, while keeping the full gain at
low density (2.4$\times$ at $m{=}1$, Appendix~\ref{app:density}). Personal haplotypes, one person's variants on one chromosome copy, are denser still, with about 1.3 variants per kilobase in the 1000 Genomes panel.
Their edits fall in the dependency intervals of most trunk bins, so in unstitched development-window tests ShiftCache gives no
speedup.

\paragraph{Execution-method selection.} Separately from view selection, an offline linear model, the \emph{method selector}, predicts campaign time
from the planned counts of full encoder passes, tiles and transformer passes. On 16 campaigns from the offset-diversity test, none used in fitting but all from the same windows, the selector picks
the fastest measured method each time, including the crossover to layer-local reuse, where always choosing ShiftCache costs up to 48\%. We have not tested it on independent windows,
small campaigns or other GPUs (Appendix~\ref{app:selector}).

\subsection{Transfer to Borzoi and Enformer}
\label{sec:enformer}
The planner needs only the locality conditions of \S\ref{sec:correctness}. We test it on Borzoi,
whose U-Net reads three encoder outputs, and on Enformer, which pools by attention.

\paragraph{Borzoi with multiple encoder outputs.} Borzoi's encoder, with the skip-connection projections we cache, accounts for 54\% of its FLOPs, so removing it would reduce total FLOPs by at most about 2.2$\times$. This is not a measured latency bound.
On the 37 to 59 real indels per held-out 524~kb window that Borzoi's stitching supports (pure insertions or deletions
inside the region Borzoi predicts), ShiftCache is 1.67$\times$ faster than \dedupdense{} (95\% interval 1.62 to 1.71). It is 1.93$\times$ faster on a deletion
scan, 1.92$\times$ on matched insertions, and 2.5 to 4.1$\times$ faster than three-pass execution (Appendix~\ref{app:borzoi}).

\paragraph{Enformer with attention pooling.} Enformer's attention-pooling weights depend only on each aligned pair, so the assumptions of \S\ref{sec:correctness} hold. An adapter checks these assumptions, derives the trunk's
dependency interval and separates the encoder, which accounts for 77.6\% of forward-pass FLOPs, while the executor and planner are unchanged. On 128 real indels per held-out window with unstitched scoring, ShiftCache takes 10.4~s against
27.1~s for dense execution, a 2.60$\times$ speedup. On 64 random indels of 1 to 31 bases, it is 2.24$\times$ faster, and aligned tiles reach 1.45$\times$ and 1.42$\times$. For every reuse method, 308 of $3.7\times10^{9}$ output values, in eight variants, exceed the tolerance, with a largest absolute difference of
$2.1\times10^{-4}$. Aligned tiles, which build no views, show the same failures, so the failures likely come from cuDNN rounding that depends on tensor shape, as on the H100. The same comparisons on the CPU pass the tolerance (Appendix~\ref{app:enformer}).

\subsection{Reimplemented Published Analyses}
\label{sec:apps}

\paragraph{Borzoi promoter deletion scans.} ShiftCache is 4.0 to
4.1$\times$ faster than our dense implementation of the three-pass protocol and 1.94$\times$ (95\% interval 1.93 to 1.95) faster than \dedupdense{}
(Table~\ref{tab:apps}). On the A6000, all
$5.3\times10^8$ HBG1 and $6.2\times10^8$ LDLR reference and alternate values pass the tolerance. ShiftCache's peak memory is slightly higher than dense execution's (7.0~GiB against 6.1 to 6.3~GiB).
On an H100, ShiftCache is 2.07 to 2.08$\times$ faster than \dedupdense{}.
Appendix~\ref{app:mpra} compares the per-base scores with measured deletion effects.

\paragraph{AlphaGenome motif ablation.} All deletions in a motif's window share one
view, $S_{-12}$. At $K{=}500$, both methods run every deletion, and ShiftCache is 3.75$\times$ faster (geometric mean, 3.73 to 3.76$\times$ per motif) with no tolerance failures, and fold changes differ from dense execution by at most
$2.1\times10^{-6}$.
On an H100, motifs 5 and 7 run 4.39$\times$ faster than dense execution. At $K{=}3000$, the 11 motifs need 33,011 deletions in total, and ShiftCache processes each motif in 26 minutes. At the study's scoring width, significance calls for H3K27ac, a mark of active regulatory DNA, agree with the published analysis for all 11 motifs at both values of $K$, and Appendix~\ref{app:motif} reports agreement for other marks and widths.

\section{Related Work}
\label{sec:related}

\paragraph{Edited-input methods and model approximation.} Yuzu~\cite{schreiber2022yuzu} reconstructs perturbed activations through
compressed sensing, and TISM~\cite{sasse2024tism} approximates in silico mutagenesis with gradients. Both were designed for substitutions,
which keep sequence length fixed. Incrementally computable transformers~\cite{sharir2023incrementally} handle token
insertions and deletions exactly, using a retrained, vector-quantized model, while ShiftCache runs the unmodified pretrained
model. AlphaGenome's distilled student model approximates its ensemble~\cite{avsec2026alphagenome}, and our reuse matches dense execution in exact arithmetic
for any encoder that meets the conditions of \S\ref{sec:correctness}.

\paragraph{Temporal and delta reuse.} DeltaCNN~\cite{parger2022deltacnn} propagates sparse frame differences through
CNNs, DeepCache~\cite{xu2018deepcache} reuses cached results for image blocks matched to a previous frame, and Eventful
Transformers~\cite{dutson2023eventful} recompute only the tokens that have changed significantly. EVA2~\cite{buckler2018eva2} warps cached activations by motion vectors and notes that shifts
off the stride grid break reuse. ShiftCache restores reuse under such shifts by encoding the shifted reference.

\paragraph{Context reuse in LLM serving.} Prefix sharing~\cite{zheng2024sglang,juravsky2024hydragen} reuses attention states only
for identical prefixes. Prompt Cache~\cite{gim2024promptcache} declares reusable prompt segments and their positions in a schema, so
reused attention states keep consistent positions. CacheBlend~\cite{yao2025cacheblend} and EPIC~\cite{hu2024epic} reuse non-prefix chunks with
approximate recovery of context-dependent states, and PatchKV~\cite{patchkv} approximately recovers the key-value cache after
context edits. ShiftCache copies matching outputs from encodings of shifted references, with equality
in exact arithmetic.

\section{Limitations}
\label{sec:limitation}

\paragraph{Workload and model scope.} Scoring each variant in its own centered 131~kb window makes ShiftCache only 1.36$\times$ faster than \dedupdense{}, because the
largest gains require shared windows and recurring offsets. When only one or two insertion requests share a length, ShiftCache is up to 1.48$\times$
slower than layer-local reuse. Personal haplotypes showed no gain over dense execution in development-window tests. For AlphaGenome, we accelerate only the forward-strand 128-base outputs. Its official gene-level RNA scoring functions use
1-base outputs from a decoder that reads activations at every encoder level, so supporting them would require views at those levels.

\paragraph{Execution and numerical scope.} We report strict fp32 because, in the tested ports, reduced precision changes
variant effects far more than reuse does.
Agreement within the tolerance holds for the default AlphaGenome and Borzoi runs on the A6000, while a small fraction of values exceeds it with larger tile batches, with Enformer or on the H100. Bitwise identity covers only the checked AlphaGenome encoder outputs with the fixed-order kernel under \texttt{torch.compile}. The layer-local and dense baselines are our
implementations, and held-out tests cover four windows on one A6000, with key runs repeated on an H100.

\section{Conclusion}
\label{sec:conclusion}

 ShiftCache shares shifted-reference encodings across indel requests and recomputes only the bins it cannot copy. For encoders that meet the conditions of \S\ref{sec:correctness}, its outputs equal those of dense execution in exact arithmetic. In the encoders and workloads we tested, reuse across indels saves time when it matches the encoder's pooling grid and when enough requests share each shifted reference. The unchanged transformer and heads then bound the end-to-end gain.

\section*{Acknowledgments}
This work was supported in part by the generosity of SoftBank Group Corp. This work was partially supported by National Institutes of Health (NIH) grant 1U54AG075931 and National Science Foundation (NSF) award 2134998 (to J.L.-M.).

\bibliography{references}
\bibliographystyle{plainnat}

\section*{AI Usage Disclosure}
LLMs (Claude Sonnet 5.5) were used solely for proofreading, checking grammar errors, result consistency across sections, and improving the readability of the manuscript. The LLMs were not involved in defining research scope and contributions, experiment design, data analysis, or interpreting the results. All results, content and scientific claims were created entirely by the authors. The authors reviewed all the text refined by LLM to ensure accuracy and originality, and take full responsibility for the integrity and accuracy of the paper.

\clearpage
\appendix
\setcounter{lemma}{0}\renewcommand{\thelemma}{A\arabic{lemma}}\renewcommand{\theHlemma}{A\arabic{lemma}}
\setcounter{proposition}{0}\renewcommand{\theproposition}{A\arabic{proposition}}\renewcommand{\theHproposition}{A\arabic{proposition}}
\section{Proofs of the Correctness Results}
\label{app:proofs}

We prove Lemma~1, Theorems~1 to~3 and Corollary~1 (Section~\ref{sec:correctness}) and derive the clean-bin conditions of Section~\ref{sec:planning}. We use the setting stated in Section~\ref{sec:correctness}. Each convolution has stride 1, dilation 1 and odd kernel width $2r{+}1$, with $r$ positions of zero padding on each side. Pointwise operations are deterministic and the same at every position. Residual additions join feature maps with the same spatial indexing, and normalization uses fixed inference-time statistics. Each pool applies one deterministic function to aligned pairs $(2j,2j{+}1)$. No operation depends on absolute position, input length or batch contents. Every input length is a multiple of the total stride $P=2^p$, where $p$ is the number of pools. All sequence coordinates, interval endpoints, shifts and indices are integers, and every output
index is in range for its input. All statements are in exact arithmetic.

\paragraph{Encoder notation.} We represent each encoder layer by a node, and each node reads only from earlier nodes. Node $v$ produces a feature map with cumulative stride $s_v$, the spacing between neighboring positions in input bases. This stride is a power of two that divides $P$. For an input of length $L$, node $v$ has
$L/s_v$ positions, and position $i$ is \emph{in range} if $0\le i<L/s_v$, which holds exactly when $0\le s_vi<L$.
Each node has offsets $(\ell_v,h_v)$, which place the ends of its dependency interval relative to the coordinate $s_vi$. They are defined recursively. The input node has $s=1$ and
$\ell=h=0$. A convolution of radius $r$ applied to node $u$ keeps the stride and sets $\ell_v=\ell_u-rs_u$ and
$h_v=h_u+rs_u$. A pointwise operation, a normalization with fixed statistics or a residual addition keeps the common stride of its
inputs and takes the smallest $\ell$ and the largest $h$ among them. A pool applied to $u$ sets $s_v=2s_u$,
$\ell_v=\ell_u$ and $h_v=h_u+s_u$. Hence $\ell_v\le0\le h_v$ for every node. The interval of position $i$ of
node $v$ is $J_v(i)=[s_vi+\ell_v,\,s_vi+h_v]$. For an encoder output tensor $b$, this is the dependency interval $I_b(j)$ of Section~\ref{sec:correctness}, and $\bar I_b(j)$ is the part of it inside the window. Every map read by a pool has $L/s_u$ positions, an even number because $s_u<P$, so pools read complete
pairs and need no padding.

\begin{lemma}[Shifted locality]
\label{lem:shifted}
Let $U$ and $U'$ be inputs of lengths $L$ and $L'$, both multiples of $P$, and let $o$ be a multiple of $P$. Let $j$ be
an in-range index of encoder output tensor $b$ for $U$, and let $j'=j-o/s_b$. Suppose that every $i\in I_b(j)$
satisfies
\begin{enumerate}
\item[(a)] $i\in[0,L)$ if and only if $i-o\in[0,L')$, and
\item[(b)] $U[i]=U'[i-o]$ whenever $i\in[0,L)$.
\end{enumerate}
Then $j'$ is in range for $U'$, and $E(U)_b[j]=E(U')_b[j']$.
\end{lemma}

\begin{proof}
We prove the following claim for every node $v$ by induction, handling each node after the nodes it reads. If $i$ is in range for $U$ and
(a) and (b) hold for every point of $J_v(i)$, then $i'=i-o/s_v$ is in range for $U'$ and $v(U)[i]=v(U')[i']$. The
offset $o/s_v$ is an integer because $s_v$ divides $P$ and $P$ divides $o$. Taking $v$ to be the node that produces
tensor $b$ gives the lemma.

\emph{Range.} Because $\ell_v\le0\le h_v$, the coordinate $s_vi$ lies in $J_v(i)$, and it lies in $[0,L)$ because $i$ is in
range. By (a), $s_vi-o=s_vi'$ lies in $[0,L')$, so $i'$ is in range for $U'$.

\emph{Input node.} Here $J_v(i)=\{i\}$, and the claim is condition (b).

\emph{Pointwise operations, normalization with fixed statistics and residual additions.} Node $v$ reads each of its inputs $u$ at the
same position $i$, and $J_u(i)\subseteq J_v(i)$ because $v$ takes the smallest $\ell$ and the largest $h$ of its inputs.
By the induction hypothesis $u(U)[i]=u(U')[i']$ for every input $u$. The operation is the same function for $U$ and
$U'$, because it does not depend on position, input length or batch contents, so $v(U)[i]=v(U')[i']$.

\emph{Convolution of radius $r$.} Node $v$ reads $u$ at the positions $m=i-r,\dots,i+r$ and uses zero for positions out
of range. For each such $m$, both the coordinate $s_um$ and the interval $J_u(m)$ lie in
$[s_ui+\ell_u-rs_u,\,s_ui+h_u+rs_u]=J_v(i)$. Applying (a) to the point $s_um$ shows that $m$ is in range for $U$ exactly
when $m'=m-o/s_u$ is in range for $U'$. If neither index is in range, both reads are zero. If both are in range, the induction hypothesis gives $u(U)[m]=u(U')[m']$. The positions $m'$ are $i'-r,\dots,i'+r$, which are the positions that $v$ reads at $i'$. The weights are shared, so $v(U)[i]=v(U')[i']$.

\emph{Pool.} Node $v$ reads $u$ at positions $2i$ and $2i+1$, both in range because $u$ has twice as many positions as
$v$. Their intervals $J_u(2i)$ and $J_u(2i+1)$ lie in $[s_vi+\ell_u,\,s_vi+s_u+h_u]=J_v(i)$, so the induction hypothesis
applies to both. Because $o$ is a multiple of $s_v=2s_u$, the corresponding positions for $U'$ are
$2i-o/s_u=2i'$ and $2i+1-o/s_u=2i'+1$. These form the aligned pair that $v$ reads at $i'$, and the pooling function is
shared, so $v(U)[i]=v(U')[i']$. This is the only step that requires the offset to be a multiple of the pooling stride.
\end{proof}

\begin{restated}{Lemma~1}{Locality}
$E(X)_b[j]$ depends only on the values and coordinates of $X$ on $\bar I_b(j)$ and on the coordinates at which the window clips $I_b(j)$.
\end{restated}

\begin{proof}
Lemma~\ref{lem:shifted} with $o=0$ shows that two inputs that agree on $\bar I_b(j)$ and satisfy condition (a) give the
same value at $j$. For two inputs of the same length, (a) holds automatically. In general, (a) states that the two
inputs clip $I_b(j)$ at the same ends and at the same coordinates.
\end{proof}

\begin{restated}{Theorem~1}{Copy validity}
Let $Y$ be a reference sequence defined on the window and the genomic flanks needed by the shifts, and let
$Y_d[i]=Y[i-d]$ for $0\le i<L$. If $X[i]=Y_d[i]$ for all $i\in\bar I_b(j)$, then $E(X)_b[j]=E(Y_d)_b[j]$ (both are
padded identically at the window edges).
\end{restated}

\begin{proof}
Apply Lemma~\ref{lem:shifted} with $U=X$, $U'=Y_d$, $L'=L$ and $o=0$. Condition (a) holds because both inputs have length
$L$, and condition (b) is the hypothesis.
\end{proof}

\begin{restated}{Theorem~2}{Phase reindexing}
Let $d=\phi+Pq$ with $q\in\mathbb{Z}$ and $0\le\phi<P$. Then $E(Y_d)_b[j]=E(Y_\phi)_b[j-(P/s_b)q]$ wherever both intervals are unclipped. For shifts not divisible by
$2^l$, the assumptions do not in general guarantee any integer reindexing of the unshifted encoding at stride $2^l$, because the pooling groups differ, although particular inputs or weights can yield coincidentally equal outputs.
\end{restated}

\begin{proof}[Proof of the first statement]
Apply Lemma~\ref{lem:shifted} with $U=Y_d$, $U'=Y_\phi$, $L'=L$ and $o=Pq$, so that $j'=j-(P/s_b)q$. The two intervals are
$I_b(j)$ and $I_b(j')=I_b(j)-Pq$. Both lie in $[0,L)$, so every $i\in I_b(j)$ has $i\in[0,L)$ and $i-Pq\in[0,L)$, which
gives (a) and shows that $j$ is in range. For (b), $Y_d[i]=Y[i-\phi-Pq]=Y_\phi[i-Pq]$.
\end{proof}

The following counterexample proves the second statement.

\begin{proposition}
\label{prop:counter}
Let $l\ge1$ and let $d$ be an integer that is not a multiple of $2^l$. There exist an encoder that satisfies the assumptions and has a stride-$2^l$ output, and a reference $Y$, with the following property. Fix any integer $m$ and consecutive indices $k,k{+}1$ such that the intervals of $E(Y_d)$ at $k,k{+}1$ and of $E(Y_0)$ at $k{-}m,k{+}1{-}m$ are all unclipped. Then $E(Y_d)[k]\ne E(Y_0)[k-m]$ or $E(Y_d)[k+1]\ne E(Y_0)[k+1-m]$.
\end{proposition}

\begin{proof}
Set $P=2^l$ and choose a window length $L=NP$ with an integer $N\ge2$. Represent each base by a one-hot vector, and let the encoder be a width-1 convolution that selects the A channel, followed by $l$ stride-2 max pools. This encoder satisfies the assumptions. Its output at index $k$ is 1 exactly when an A occurs in the input positions $[2^lk,\,2^lk+2^l)$, and its interval is $[2^lk,\,2^lk+2^l-1]$. Let $Y$ be periodic with period $2^{l+1}$, so that position $i$
holds A if $i \bmod 2^{l+1}<2^l$ and C otherwise, on the window and on its flanks. For each integer $c$, the block $[2^lc,\,2^lc+2^l)$ is then all A when $c$ is even and all C when $c$ is odd. So $E(Y_0)$ alternates between 1 and 0 at consecutive indices. For
$Y_d$, index $k$ covers the reference positions $[2^lk-d,\,2^lk-d+2^l)$. This interval has length $2^l$ and does not start at a multiple of $2^l$, so it meets two consecutive blocks, one of A and one of C. Hence $E(Y_d)[k]=1$ for every index $k$ whose interval is unclipped. At two consecutive such indices, $E(Y_d)$ equals $(1,1)$. At any two consecutive indices, $E(Y_0)$ equals $(1,0)$ or $(0,1)$. The reindexing by $m$ therefore fails at $k$ or at $k+1$.
\end{proof}

The construction uses max pooling, as in AlphaGenome and Borzoi. Particular inputs, such as a constant sequence, can
still give equal outputs by coincidence.

\begin{restated}{Theorem~3}{Tile exactness}
Let $t\ge0$ be an integer and $T$ a positive multiple of $P$ with $tP+T\le L$, and let $X_t=X[tP,\,tP+T)$ be the
corresponding tile.
Then $E(X_t)_b[j-tP/s_b]=E(X)_b[j]$ whenever $\bar I_b(j)$ lies inside the tile, with
$tP=0$ if $s_bj+\ell_b<0$ and $tP+T=L$ if $s_bj+h_b\ge L$. Thus, a dependency interval may cross a tile
boundary only where that boundary coincides with the corresponding window edge.
\end{restated}

\begin{proof}
Apply Lemma~\ref{lem:shifted} with $U=X$, $U'=X_t$, $L'=T$ and $o=tP$, so that $j'=j-tP/s_b$. Condition (b) holds
because $X[i]=X_t[i-tP]$ for every $i$ in the tile, and $\bar I_b(j)$ lies in the tile. For condition (a), take
$i\in I_b(j)$. If $i\in[0,L)$, then $i\in\bar I_b(j)$, which lies in $[tP,tP+T)$, so $i-tP\in[0,T)$. If $i<0$, then
$s_bj+\ell_b<0$, so $tP=0$ and $i-tP<0$. If $i\ge L$, then $s_bj+h_b\ge L$, so $tP+T=L$ and $i-tP\ge T$. In each case
$i\in[0,L)$ exactly when $i-tP\in[0,T)$.
\end{proof}

The side conditions prevent a tile from using zero padding where the full window has sequence values. They require a
dependency interval to leave the tile only at a window edge, where the tile and the window both pad with zeros.

\begin{proposition}[Clean-bin conditions]
\label{prop:clean}
Let $[a,e)$ be a segment with $0\le a<e\le L$, and let $j$ be an in-range index of tensor $b$, with
$\ell=\ell_b$, $h=h_b$ and $s=s_b$. Then $\bar I_b(j)\subseteq[a,e)$ if and only if
$j\ge\lceil(a-\ell)/s\rceil$ when $a>0$ and $j\le\lfloor(e-1-h)/s\rfloor$ when $e<L$.
\end{proposition}

\begin{proof}
The clipped interval is $[\max(0,sj+\ell),\,\min(L-1,sj+h)]$, which is nonempty because it contains $sj$. It lies in
$[a,e)$ exactly when $\max(0,sj+\ell)\ge a$ and $\min(L-1,sj+h)\le e-1$. If $a=0$, the first condition always holds.
If $a>0$, it holds exactly when $sj+\ell\ge a$, that is, when $j\ge\lceil(a-\ell)/s\rceil$, since $j$ is an integer. If
$e=L$, the second condition always holds. If $e<L$, it holds exactly when $sj+h\le e-1$, that is, when
$j\le\lfloor(e-1-h)/s\rfloor$.
\end{proof}

For a segment with offset $d$, every input position $i$ in the segment satisfies $X[i]=\mathrm{ref}[i-d]=S_d[i]$. By Theorem~1, a bin $j$ of $E(X)$ that meets these conditions equals bin $j$ of $E(S_d)$.

\begin{restatednonote}{Corollary~1}
In exact arithmetic, assembling every bin according to Theorems~1 to~3 yields $E(X)$.
Applying the same tail, stitching and heads then gives the dense outputs.
\end{restatednonote}

\begin{proof}
The planner assigns each bin $j$ of each encoder output tensor one source. The reference is the view with $d=0$. A request that falls back to a full pass computes $E(X)$ directly, which is the tile case $t=0$, $T=L$ of Theorem~3. A partial view built from tiles keeps only bins that satisfy Theorem~3 for its shifted reference, so those
bins equal the corresponding bins of $E(S_d)$. A copied bin satisfies the conditions of
Proposition~\ref{prop:clean} for a segment with offset $d$ and is read from $E(S_d)$, so it equals $E(X)_b[j]$ by
Theorem~1. A bin read from a phase view $E(S_\phi)$ at index $j-(P/s_b)q$ has both intervals unclipped, so it equals $E(S_d)_b[j]$ by Theorem~2. It therefore equals $E(X)_b[j]$ by Theorem~1. Boundary bins are recomputed. A recomputed bin is
kept only from a tile that satisfies the conditions of Theorem~3, so it equals $E(X)_b[j]$. Every bin of every tensor
therefore equals its dense value, and the assembled tensors equal $E(X)$. For the outputs we evaluate, the
downstream model reads sequence values only through these encoder output tensors, including all three tensors for
Borzoi. Given the same edit endpoints and scoring-protocol settings, the downstream model and stitching are
deterministic, so applying them in the order the protocol prescribes produces the dense outputs.
\end{proof}

\section{Workloads and Protocol Details}
\label{app:protocol}

\paragraph{Genomic windows.} We use the hg38 human reference genome throughout. Design and tuning used eight earlier windows on chromosomes 1, 17 and 19. Four of them were development windows, and four were initially held out. After ShiftCache's settings were frozen, we used a fresh seed (20261003) to draw the four held-out windows reported here. Their 524~kb windows start at chr2:7,830,684, chr6:163,961,849, chr11:123,986,934 and chr12:932,447 (0-based). None of them overlaps an earlier window. We added the layer-local baseline after the first held-out evaluation and later redesigned it to reuse more work. Its tuning used development windows (Appendix~\ref{app:lltune}). Each window contains 524,288 bases plus 4,096 bases of genomic flank on each side, and the 131,072-base window is its central part. Application windows are centered on the published sites and follow the published code's coordinate convention. The promoter windows have 524~kb and are centered on HBG1 (chr11:5,249,805 to 5,250,078) and LDLR (chr19:11,089,231 to 11,089,548). These coordinates are 1-based and inclusive. Each TSC22D3 CEBPB motif window has $2^{20}$ bases.

\paragraph{Real indels.} We use the high-coverage 1000 Genomes panel, which assigns variants to chromosome copies~\cite{byrska2022kgp}. We take every indel of at most 50 bases that has one alternate allele and whose reference span lies in the window. Structural variants are excluded. We check each reference allele against the reference genome and trim the bases that the reference and alternate alleles share. We then draw a seeded sample (seed 20261002) of 256 indels per 131~kb
window or 128 per 524~kb window. Borzoi's baskerville stitching needs an insertion with no deleted bases or a deletion with no inserted bases, inside the central region Borzoi predicts. This leaves 37 to 59 indels per held-out window.

\paragraph{Timing and validation protocol.} For the warmed measurements, taken after compilation, a timed campaign starts with deduplication and ends when all reference and alternate predictions are in CPU memory. Before these measurements, each process compiles every tensor shape and calibrates the cost table. First-use measurements include these costs as specified in Appendix~\ref{app:tuning}. Timing excludes the final joining of output chunks in CPU memory, except in the motif runs. Repetitions rotate the order of the methods. The promoter applications repeat the \dedupdense{} and ShiftCache measurements. \threepass{}, the motif runs and the timed exhaustive scans are measured once. The centered-window, offset-diversity and TF32 runs use two campaigns at 131~kb, and the promoter scans use three at 524~kb. Agreement with the run's dense baseline is checked on the first
repetition. The timed exhaustive scans and the first-use runs make no such check, and separate runs check the exhaustive scans' full outputs (Section~\ref{sec:exact}). The motif runs compare the MM.1S tracks they score, and at $K{=}3000$ only the first 64 deletions per motif. An output value passes the tolerance if it differs from the dense value by at most
$10^{-5}+10^{-5}|\mathrm{dense}|$ (Section~\ref{sec:eval}). The CPU tests with random weights compare every execution method with dense execution for substitutions, insertions, deletions and haplotypes (variants on one chromosome copy). The cases include edits at window edges, shifts divisible by 128, and an insertion and a deletion whose lengths cancel. The tests also cover centered windows and the stitching conventions of AlphaGenome and baskerville. We implement layer-local reuse only for AlphaGenome, and it is tested with stitching only in shared windows.

\section{Dense Baseline Tuning and First-Use Costs}
\label{app:tuning}
We tuned \dedupdense{}'s sequence batch size (1 to 16), the number of requests per stitching chunk (16 to 128) and its compilation settings (maximum autotuning and no compilation). The best setting reduces campaign time by at most 3\% compared with the default (Table~\ref{tab:tuning}).

\begin{table}[H]
\centering
\caption{Seconds per campaign for \dedupdense{} and ShiftCache on the first development window at 131~kb, as the median of three campaigns (RTX A6000, strict fp32). The table shows selected settings of the sweep. Batch is the number of sequences per full pass, and chunk is the number of requests stitched together.}
\label{tab:tuning}
\small
\begin{tabular}{lrrrr}
\toprule
& \multicolumn{2}{c}{256 real indels} & \multicolumn{2}{c}{512 deletions} \\
Setting & \dedupdense & ShiftCache & \dedupdense & ShiftCache \\
\midrule
Default (batch 8, chunk 32) & 82.8 & 17.2 & 84.4 & 14.3 \\
Batch 1 & 83.3 & 16.9 & 82.9 & 13.7 \\
Batch 2 & 83.9 & 17.4 & 83.4 & 14.1 \\
Chunk 16 & 83.6 & 17.2 & 88.1 & 14.8 \\
Chunk 128 & 81.4 & 17.0 & 82.0 & 14.1 \\
Max-autotune compilation & 82.4 & 16.8 & 83.9 & 14.0 \\
Uncompiled & 98.9 & 23.1 & 100.5 & 19.8 \\
\bottomrule
\end{tabular}
\end{table}

\begin{table}[H]
\centering
\caption{Each first-use run starts in a fresh process with empty compilation caches. It scores 256 real indels in the first held-out window at 131~kb (RTX A6000, strict fp32). Times are seconds per campaign and exclude model loading, which takes about 7~s.}
\small
\begin{tabular}{lrr}
\toprule
Method & First campaign & Later campaigns \\
\midrule
\threepass & 238 & n/a \\
\dedupdense & 222 & 78 \\
ShiftCache, saved cost table & 194 & 17 \\
ShiftCache, calibrating in process & 347 & 17 \\
\bottomrule
\end{tabular}
\end{table}

\section{Numerical Agreement and Memory Use}
\label{app:numerics}

On two development windows, our baseline is compiled strict-fp32 dense execution. We compare it with dense execution at batch size 1, dense execution without compilation (eager mode), separate encoder and transformer runs, aligned tiles and ShiftCache. These unstitched runs use 16 substitutions, 16 indels and 16 haplotypes per window. A variant's effect is its alternate prediction minus its reference prediction. Its relative error is the $L_2$ norm of the difference from the baseline effect, divided by the $L_2$ norm of the baseline effect. We report medians over variants. On the A6000, the reuse methods have median relative errors in substitution effects of $4.9\times10^{-4}$ and $5.9\times10^{-4}$ on the two windows. These errors are below the differences between dense runs at different batch sizes ($1.7\times10^{-3}$ and $3.3\times10^{-3}$) and between compiled and eager dense execution ($1.3\times10^{-3}$ and $1.4\times10^{-3}$). On the H100, ShiftCache's median relative effect error ranges from $0.7\times10^{-3}$ to
$1.0\times10^{-3}$ for substitutions and from $4.0\times10^{-5}$ to $7.8\times10^{-5}$ for indels. The differences between dense runs on the H100 range from $1.2\times10^{-3}$ to $3.4\times10^{-3}$ for substitutions and from $2.5\times10^{-5}$ to $1.1\times10^{-4}$ for indels. In one diagnostic run, cuDNN's deterministic mode does not remove the H100 tolerance failures. Reduced precision produces much larger errors for every method. With TF32 on the A6000, median relative effect errors are 40 to 71\% for substitutions, 0.8 to 2.4\% for indels and 2.5 to 3.8\% for haplotypes. These errors are similar for dense execution and reuse. Automatic mixed precision with bf16 gives larger errors in this port, so we report strict fp32. With TF32 enabled, ShiftCache's held-out
speedups over \dedupdense{} at 131~kb are 4.01$\times$ on real indels, 4.74$\times$ on the deletion scan and 5.46$\times$ on matched insertions.

Measured against \threepass{} on the held-out AlphaGenome campaigns of Table~\ref{tab:main}, no ShiftCache request has a value outside the tolerance on the A6000 (0 of 5,376 requests). At 131~kb on the H100, 402 of 3,584 ShiftCache requests have at least one such value, and 370 of these are matched insertions. \dedupdense{} has no such request on either GPU.

On the 11,008-request exhaustive scan, ShiftCache caches the outputs of at most 273 distinct sequences at once. In the warmed campaigns at 131~kb, layer-local reuse uses 7.1 to 9.9~GiB of peak allocated GPU tensor memory. ShiftCache uses less of this memory than \dedupdense{} in the AlphaGenome campaigns. It uses 7.8 to 11.8~GiB against 13.0 to 15.7~GiB at 131~kb and 13.0 to 17.6~GiB against 16.0 to 19.6~GiB at 524~kb. It uses 14.1~GiB against 22.1~GiB on the exhaustive scan and 10.8~GiB against 11.3~GiB at 1~Mb. In the Borzoi promoter scans ShiftCache uses slightly more than \dedupdense{}, 7.0~GiB against 6.1~GiB. The campaigns in these memory measurements used 1 to 30 views. These measurements exclude memory reserved by the allocator and do not establish the GPU capacity
required to run each method in isolation.

\begin{figure}[t]
\centering
\includegraphics[width=\textwidth]{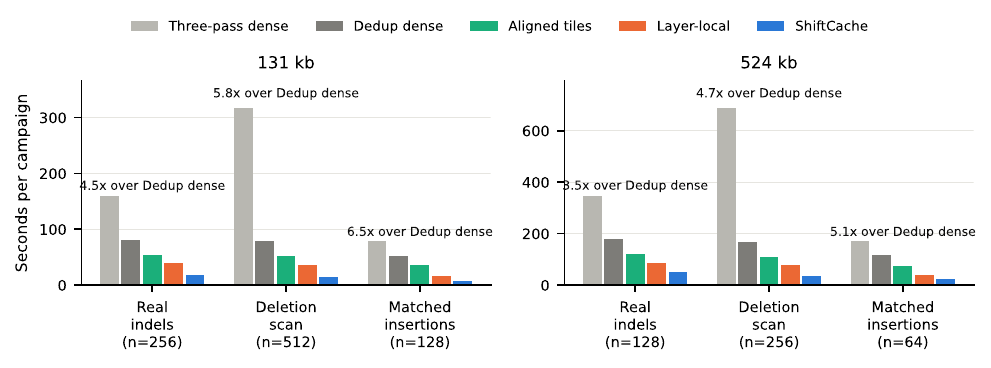}
\caption{Seconds per stitched campaign on four held-out windows (geometric mean), with AlphaGenome in fp32 on one RTX A6000. Here $n$ is the number of requests. Labels give ShiftCache's speedup over \dedupdense{}, and Table~\ref{tab:main} gives the 95\% intervals of these speedups. \threepass{} was measured separately on the same windows, and all other methods were
measured together.}
\label{fig:results}
\end{figure}

\section{Bitwise-Mode Kernel and Costs}
\label{app:bitwise}
Our GPU kernel, written in Triton, computes one-dimensional convolution by matrix multiplication on small blocks. It does not build the larger input array that would repeat values for each kernel position. Every tensor shape uses blocks of 128 positions and 64 output channels, with 16 input channels processed per step. The kernel runs on four warps (groups of 32 GPU threads) with three pipeline stages. Autotuning is disabled, and each output sum stays within one thread block. The kernel visits convolution kernel positions in an outer loop and input-channel blocks in an inner loop, both in a fixed order. It accumulates full fp32 (non-TF32) dot products into one fp32 output block and adds the bias once after both loops. Masked loads, which read only valid entries, handle zero padding and incomplete blocks. AlphaGenome's standardized convolutions recompute their standardized weights on every call. Bitwise mode computes them once when it is enabled, so runs with and without compilation use identical weights. Other operations still round differently between
eager and compiled execution, so we check bitwise agreement against compiled dense execution with the same kernel.

On each GPU, the bitwise check covers 34 assembled encoder outputs and 17 views. The outputs are deletion alternates and their right-matched references at 131~kb and 1~Mb. In this mode, the largest remaining absolute differences in the reuse methods' final outputs are $5.3\times10^{-4}$ in the held-out campaigns and $3.2\times10^{-3}$ in the checked 1~Mb motif outputs. All of these pass the tolerance, whose relative term grows with the magnitude of each dense value. The differences arise after the encoder. Changing the transformer batch size from 1 to 8 changes a sequence's embeddings by up to $7.6\times10^{-6}$, for dense execution as much as for reuse.

On the A6000, the fixed kernel multiplies \dedupdense{} runtime by 1.73 to 1.75 and ShiftCache runtime by 1.29 to 1.44. On the H100, these factors are 1.58 to 1.59 and 1.18 to 1.32. With both methods using the fixed kernel, ShiftCache is 5.4 to 8.7$\times$ faster than \dedupdense{}. Compared with default \dedupdense{}, ShiftCache with the fixed kernel is 3.15 to 4.97$\times$ faster on the A6000 and 3.42 to 5.49$\times$ faster on the H100.

\section{Layer-Local Reuse Tuning}
\label{app:lltune}

\begin{table}[H]
\centering
\caption{Campaign time relative to default layer-local reuse, on two development windows at 131~kb (geometric mean over windows). Each row gives the small and large tile sizes in trunk bins, then the tile batch. The default uses 32-bin and 128-bin tiles with a tile batch of 16. With tile batches above 16, outputs fail the tolerance against \dedupdense{}. The last column gives the fraction of reference and alternate output values that fail, pooled over both windows and all three workloads. ShiftCache keeps a tile batch of 16 for the same reason.}
\small
\begin{tabular}{lrrrr}
\toprule
Tile sizes, tile batch & Real indels & Deletion scan & Matched insertions & Failing fraction \\
\midrule
32/128, 16 (default) & 1.00 (40.3 s) & 1.00 (36.5 s) & 1.00 (16.6 s) & 0 \\
32/128, 8 & 1.02 & 1.01 & 1.02 & 0 \\
16/64, 16 & 1.01 & 0.99 & 0.95 & 0 \\
16/128, 16 & 0.98 & 0.97 & 0.94 & 0 \\
64/256, 8 & 1.09 & 1.19 & 1.23 & 0 \\
64/256, 16 & 1.08 & 1.18 & 1.21 & 0 \\
16/64, 32 & 0.95 & 0.94 & 0.94 & $2.9\times10^{-4}$ \\
32/128, 32 & 0.91 & 0.92 & 0.96 & $3.6\times10^{-4}$ \\
32/128, 64 & 0.65 & 0.66 & 0.71 & $3.6\times10^{-4}$ \\
ShiftCache (default) & 0.42 & 0.37 & 0.49 & 0 \\
\bottomrule
\end{tabular}
\end{table}

Among settings that pass the tolerance, development tuning reduces campaign time by at most 6.4\% compared with default layer-local reuse. Larger tile batches speed up both methods, but their outputs fail the tolerance, likely because cuDNN selects different algorithms for those tensor shapes. With both methods at a tile batch of 64 on the held-out windows, layer-local reuse takes 25.7, 24.2 and 12.0~s on real indels, the deletion scan and matched insertions, and ShiftCache takes 14.4, 12.3 and 7.5~s, measured separately on the same windows. ShiftCache at its default tile batch of 16 takes 18.3, 13.8 and 8.2~s when measured together with layer-local reuse at a tile batch of 64. At a tile batch of 64, the fraction of values that exceed the tolerance is $2.6\times10^{-3}$ for ShiftCache and $4.1\times10^{-4}$ for layer-local reuse. With both methods at equal tile batches, ShiftCache is 1.60 to 1.98$\times$ faster than layer-local reuse. ShiftCache at its default tile batch of 16 is 1.41 to 1.76$\times$ faster than layer-local reuse at a tile batch of 64.

\section{Sensitivity to Execution Settings}
\label{app:ablation}

\begin{table}[H]
\centering
\caption{ShiftCache speedup over \dedupdense{} on the first development window at 131~kb, with static view selection. The last row gives the time of greedy view selection divided by the time of static view selection, so a smaller value is better. Tile sizes are in trunk bins.}
\small
\begin{tabular}{lrr}
\toprule
Configuration & Real indels & Deletion scan \\
\midrule
Default (tiles 32/128, partial views) & 4.83 & 5.88 \\
Transformer batch 4 & 4.90 & 5.96 \\
Transformer batch 16 & 4.89 & 5.90 \\
Tiles 16/64 & 4.33 & 5.10 \\
Tiles 64/256 & 4.30 & 5.09 \\
Full-length views only & 4.93 & 5.88 \\
Greedy view selection (time ratio) & 1.02 & 1.00 \\
\bottomrule
\end{tabular}
\end{table}

On the development windows, greedy view selection takes 1.00 to 1.03 times as long as the static rule, so we fixed the static rule as the default before the held-out evaluation. On the held-out shared-window campaigns, with one indel per alternate, greedy view selection takes 1.00 to 1.04 times as long, and on per-variant centered 131~kb windows it takes 1.42 times as long.

\paragraph{Tile coverage.} In the held-out AlphaGenome campaigns at 131~kb, when all required views are available, fewer than 1\% of alternate trunk bins are dirty, meaning that they must be recomputed. Tiles, including their halos of extra context, cover 3.2 to 3.9\% of all alternate trunk bins in the real-indel campaigns. Full-pass fallbacks and view construction add encoder work. The runtime breakdown in Section~\ref{sec:breakdown} includes all of these costs.

\paragraph{Second GPU.} On an NVIDIA H100 at 131~kb, ShiftCache is faster than \dedupdense{} by 4.51$\times$ on real indels (95\% interval 4.31 to 4.73), 5.65$\times$ on the deletion scan (95\% interval 5.55 to 5.75) and 6.50$\times$ on matched insertions. The same speedups on the A6000 are 4.52$\times$, 5.81$\times$ and 6.47$\times$. \dedupdense{}, aligned tiles and ShiftCache each run about 2.2$\times$ faster on the H100 than on the A6000. We did not run the reported layer-local baseline on the H100.

\section{Edit Density and Placement}
\label{app:density}

\begin{table}[H]
\centering
\caption{Seconds per unstitched campaign of 32 sequences, each with $m$ random indels of 1 to 4 bases (geometric mean of two held-out windows at 131~kb). Clustered edits lie within 4~kb, and dispersed edits are spread across the window. The table shows selected clustered densities and every tested dispersed density. All comparisons with dense execution pass the tolerance. The last column gives the number of views built per window by each rule.}
\label{tab:density}
\small
\begin{adjustbox}{max width=\textwidth}
\begin{tabular}{llrrrrl}
\toprule
Layout & $m$ & Dense & Aligned tiles & ShiftCache, static & ShiftCache, greedy & Views (static / greedy) \\
\midrule
Clustered (4 kb) & 1 & 6.63 & 4.57 & 2.44 & 2.45 & 7-8 / 7-8 \\
 & 4 & 6.63 & 4.67 & 4.05 & 4.11 & 12-13 / 2-9 \\
 & 16 & 6.63 & 4.83 & 4.13 & 4.15 & 4-6 / 2-4 \\
 & 64 & 6.64 & 5.11 & 4.94 & 5.13 & 2-6 / 0 \\
Dispersed & 1 & 6.62 & 4.56 & 2.77 & 2.77 & 7-8 / 7 \\
 & 2 & 6.63 & 5.18 & 3.54 & 3.98 & 10-11 / 6-7 \\
 & 4 & 6.63 & 6.21 & 4.41 & 4.59 & 14-15 / 6-9 \\
 & 8 & 6.63 & 6.72 & 6.08 & 6.46 & 21-22 / 2 \\
 & 16 & 6.63 & 7.21 & 9.41 & 7.05 & 31-40 / 1-2 \\
 & 32 & 6.64 & 6.84 & 13.14 & 6.93 & 46-55 / 0 \\
 & 64 & 6.64 & 6.85 & 6.85 & 7.00 & 0 / 0 \\
\bottomrule
\end{tabular}
\end{adjustbox}
\end{table}

\section{Transfer to Borzoi}
\label{app:borzoi}

\begin{table}[H]
\centering
\caption{ShiftCache speedups for Borzoi on held-out 524~kb windows with baskerville's stitching convention. Output comparisons use each track's sum over positions. The 95\% interval is reported for real indels.}
\small
\begin{tabular}{lrr}
\toprule
Workload & Over \dedupdense{} [95\%] & Over \threepass{} \\
\midrule
Real indels (37 to 59 per window) & \textbf{1.67} [1.62, 1.71] & 2.5 \\
Deletion scan (256) & \textbf{1.93} & 4.1 \\
Matched insertions (64) & \textbf{1.92} & 2.9 \\
\bottomrule
\end{tabular}
\end{table}

Borzoi's encoder alone accounts for 52\% of its FLOPs. The projections on the U-Net's skip connections, which we also cache, raise this share to 54\%. With baskerville's convention, Borzoi stitches 32-base predictions. Our adapter follows this convention and fills the positions shifted in at the window edge with genomic bases. Baskerville fills them with N, the symbol for an unknown base. Dense execution and reuse receive identical sequences in our comparisons. The promoter scans of \citet{korsakova2025shift} instead stitch the reference for deletions. For $n$ single-base deletions in one window, reference stitching needs at most $n{+}2$ distinct sequences and alternate stitching needs at most $2n{+}1$. Our promoter scans use alternate stitching, with 385 distinct sequences for HBG1 and 457 for LDLR, and we did not time Borzoi with reference stitching. The planner uses the dependency intervals of the three encoder outputs, at 128-base, 64-base and 32-base resolution, to construct one shared plan on the 128-base trunk grid.

\section{Execution-Method Selection}
\label{app:selector}
The method selector, an offline linear model, uses counts that the planner knows before GPU execution. These are the numbers of full encoder passes, small tiles, large tiles and transformer passes. Layer-local tiles are weighted by the approximate encoder FLOPs remaining after the last reused level. We fit the model on the held-out 131~kb campaigns of Table~\ref{tab:main}. Its median training error is 1.3\% for dense execution, aligned tiles and ShiftCache together and 1.0\% for layer-local reuse. We then test it on 16 offset-diversity campaigns from the same windows, none of which was used in fitting. It selects the fastest measured method among dense execution, aligned tiles, layer-local reuse and ShiftCache on all 16 campaigns, including those where layer-local reuse becomes fastest. Always choosing ShiftCache on these campaigns takes 7.9\% longer than the fastest method on average and 48\% longer at worst. The selector's median prediction errors on this sweep are 0.5 to 3.9\% for dense execution, aligned tiles and ShiftCache, and 5.6\% for layer-local reuse.

The edit-density study does not include layer-local reuse. There we fit one constant for transformer and head time on a single configuration, and we exclude that configuration from the evaluation. Selecting a method for each campaign is 1.4\% slower than the best measured method on average and 14\% at worst. Always using static view selection is 13\% slower on average and 98\% at worst, and always using greedy view selection is 3.8\% and 17\% slower. The selector's median prediction error for ShiftCache with static view selection is 12.8\% in this study.
We have not tested these fitted coefficients on independent windows, small campaigns or other GPUs.

\section{Transfer to Enformer}
\label{app:enformer}
The Enformer port predicts 896 output bins for 5,313 human tracks. The adapter derives the dependency interval of trunk bin $j$ as $[128j-259,\,128j+386]$, which gives a three-bin halo on each side. Counting convolution, linear and attention FLOPs, the encoder accounts for 77.6\% of a forward pass. Removing all encoder work would therefore reduce total FLOPs by at most 4.5$\times$. We use
the four held-out windows at 196,608 bases, each centered in its 524~kb window, with one alternate per indel and three
repetitions. In the real-indel campaigns ShiftCache uses 7.7~GiB of peak allocated memory against 9.9 to 12.2~GiB for
dense execution. For every reuse method, 308 of $3.7\times10^{9}$ output values exceed the tolerance against dense execution. These values occur in eight variants, with matching aggregate statistics across methods. The largest absolute difference is $2.1\times10^{-4}$. The checked CPU comparisons pass the
tolerance, and float64 tiles and full-window encodings agree up to rounding error.

\section{Motif-Ablation Agreement with Published Results}
\label{app:motif}
Here $K$ is the number of control segments per motif (Section~\ref{sec:eval}). We follow each published table's sign convention~\cite{motifablation2026}, which is $\log_2$(reference\slash ablated) at $K{=}500$ and $\log_2$(ablated\slash reference) at $K{=}3000$. The published tables were computed with DeepMind's API model. Our fold changes differ from them by about 0.2\% at the median, a median absolute difference of 0.0028 in log$_2$ fold change. Section~\ref{sec:apps} reports agreement between ShiftCache and local dense execution. The study scores fold changes over windows of 5, 7, 9, 11 and 13 bins of 128 bases centered on the motif. At its primary width of 9 bins, motifs 5 and 7 are significant for H3K27ac at $K{=}500$ and motifs 3, 5 and 7 at $K{=}3000$, as in the published tables. Across nine histone marks and the five widths, 486 of 495 significance calls agree with the published tables at $K{=}500$ and 482 of 495 at $K{=}3000$. Every disagreement lies within 0.0073, in log$_2$ fold change, of a 2.5th or 97.5th percentile bound.

\section{Comparison with MPRA Measurements}
\label{app:mpra}
\begin{figure}[H]
\centering
\includegraphics[width=0.5\textwidth]{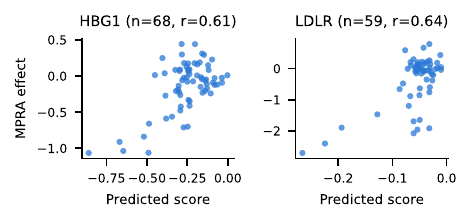}
\caption{Per-base deletion scores from ShiftCache's outputs against MPRA-measured deletion effects, for positions measured with at least 10 barcoded reporter copies. Each score is the mean of $\log_2(A_t{+}1)-\log_2(R_t{+}1)$ over the tracks that the published protocol selects, which are K562 tracks for HBG1 and adrenal gland tracks for LDLR. For track $t$, $A_t$ and $R_t$ are the alternate and reference signals summed over the central 4~kb, using only bins that overlap the gene's exons for RNA tracks. A negative score means that the deletion lowers predicted activity. Each point is one deleted base, $r$ is the Pearson correlation, and the two panels use different axis ranges.}
\label{fig:mpra}
\end{figure}

The promoter scans give one score per deleted base, using the published track selection and scoring formula with alternate stitching (Figure~\ref{fig:mpra}). The published scan stitches the reference and differs in the other ways listed in Section~\ref{sec:eval}. It also averages forward and reverse-complement predictions, and our exon annotations come from Ensembl (GRCh38), retrieved in October 2026. These scores therefore do not reproduce its values. To check that these scores follow measured effects, we compare them with deletion effects from a massively parallel reporter assay (MPRA)~\cite{kircher2019satmut}. This assay measures how strongly each tested sequence drives a reporter gene. For the 68 HBG1 positions measured with at least 10 barcoded reporter copies, the Pearson correlation is 0.61 and the Spearman correlation is 0.32. For the 59 LDLR positions, these correlations are 0.64 and 0.29. Scores from dense execution and ShiftCache agree within the tolerance, so these correlations describe the model and our scoring protocol.

\section{Reproducibility}
\label{app:repro}
We reran the held-out campaigns at both context lengths, the Borzoi promoter scans, two motifs of the motif scan and the Enformer campaigns with the same code. Each rerun reproduced its reported speedups within 0.3\%, except \threepass{}, timed once, which reproduced within 1.8\%. The numerical agreement outcomes were the same. Table~\ref{tab:main}, the offset-diversity sweep and the method selector use the layer-local implementation of Section~\ref{sec:eval}, which caches activations before each pool. The second-GPU runs use the same package versions, model weights and input windows as the A6000 runs, verified by checksum.

\end{document}